\documentclass[journal]{IEEEtran}

\usepackage{amsthm}
\newif\ifieee
\newif\ifieeeplots
\newlength{\figw}\newlength{\figwide}
\AtBeginDocument{\setlength{\figw}{\figscale\columnwidth}\setlength{\figwide}{\figscale\textwidth}}
\providecommand{\figscale}{1}
\newsavebox{\diagboxa}\newsavebox{\diagboxb}
\usepackage[T1]{fontenc}
\usepackage[utf8]{inputenc}
\usepackage{amsmath,amssymb,bm}
\usepackage{booktabs}
\usepackage{tabularx}
\usepackage{threeparttable}
\usepackage[table]{xcolor}
\usepackage{graphicx}
\usepackage{tikz}
\usetikzlibrary{arrows.meta,calc,positioning,patterns,fit,backgrounds}
\usepackage{tikz-3dplot}
\usepackage{pgfplots}
\usepackage{algorithm}
\usepackage{algpseudocode}
\usepackage{microtype}

\newcommand{\vect}[1]{\bm{#1}}
\newcommand{\mat}[1]{\bm{#1}}
\definecolor{okgreen}{RGB}{0,128,60}
\definecolor{refgray}{rgb}{0.45,0.45,0.45}
\definecolor{plancyan}{rgb}{0.30,0.75,0.93}
\definecolor{ballorange}{rgb}{0.85,0.33,0.10}
\definecolor{limitred}{rgb}{0.80,0.10,0.10}
\definecolor{ctrlblue}{rgb}{0.00,0.45,0.74}
\definecolor{ctrlyellow}{rgb}{0.93,0.69,0.13}
\definecolor{ctrlgreen}{rgb}{0.47,0.67,0.19}
\definecolor{cvred}{rgb}{0.85,0.15,0.15}
\definecolor{cpgreen}{rgb}{0.00,0.55,0.30}
\definecolor{obsbrown}{rgb}{0.60,0.35,0.10}
\definecolor{badred}{RGB}{190,30,30}
\definecolor{rowhl}{RGB}{232,240,252}
\definecolor{cellbad}{RGB}{252,226,226}
\newcommand{\yes}{\textcolor{okgreen}{$\checkmark$}}
\newcommand{\no}{\textcolor{badred}{$\times$}}
\newcommand{\nr}{\textcolor{gray}{--}}
\newcommand{\bad}[1]{\textcolor{badred}{#1}}

\newcommand{\tabgroup}[2]{\multicolumn{#1}{@{}l}{\textbf{\textit{#2}}}\\}
\newcommand{\hd}[1]{{\bfseries\boldmath #1}}
\usepackage{siunitx}
\usepackage{hyperref}
\hypersetup{colorlinks=false, pdfborder={0 0 1},
  linkbordercolor={1 0 0}, citebordercolor={0 0.7 0}, urlbordercolor={0 0.7 0.9}}
\usepackage[capitalise,nameinlink]{cleveref}
\crefname{figure}{Fig.}{Figs.}
\crefname{table}{Table}{Tables}
\crefname{section}{Section}{Sections}
\crefname{lemma}{Lemma}{Lemmas}
\crefname{proposition}{Proposition}{Propositions}
\crefname{remark}{Remark}{Remarks}
\crefname{assumption}{Assumption}{Assumptions}
\crefname{algorithm}{Algorithm}{Algorithms}

\theoremstyle{plain}
\newtheorem{proposition}{Proposition}

\newtheorem{remarkbase}{Remark}
\theoremstyle{definition}
\newtheorem{assumption}{Assumption}
\crefname{remarkbase}{Remark}{Remarks}
\NewDocumentEnvironment{remark}{o}
  {\IfNoValueTF{#1}{\remarkbase}{\remarkbase[#1]}%
   \pushQED{\qed}}
  {\popQED\endremarkbase}

\ieeetrue\ieeeplotstrue
\renewcommand{\figscale}{0.8}
\makeatletter
\def\section{\@startsection{section}{1}{\z@}{1.6ex plus 0.5ex minus 0.5ex}%
{0.5ex plus 0.3ex minus 0ex}{\normalfont\normalsize\centering\scshape}}%
\def\subsection{\@startsection{subsection}{2}{\z@}{1.4ex plus 0.5ex minus 0.5ex}%
{0.4ex plus 0.3ex minus 0ex}{\normalfont\normalsize\itshape}}%
\makeatother
\makeatletter\def\@IEEEfigurecaptionsepspace{\vskip -2pt\relax}\makeatother  

\newcommand{\firstword}[2]{\IEEEPARstart{#1}{#2}}
\newcommand{\hookwidth}{\linewidth}

\title{\textbf{\textit{Safe-Ballbot}}: Tilt-Consistent MPC with Control Barrier Functions for Ballbot Navigation}
\author{Duc~Cuong~Vu and Tung~Lam~Nguyen%
\thanks{Corresponding author: Tung Lam Nguyen.}%
\thanks{Duc~Cuong~Vu is with the School of Electrical and Electronic Engineering,
Nanyang Technological University, Singapore
(e-mail: \texttt{duccuong001@e.ntu.edu.sg}).}%
\thanks{Tung~Lam~Nguyen is with the Motion Control and Applied
Robotics Laboratory (MoCAR), School of Electrical and Electronic Engineering, Hanoi
University of Science and Technology, Hanoi, Vietnam
(e-mail: \texttt{lam.nguyentung@hust.edu.vn}).}}

\begin{document}

\maketitle

\begin{abstract}
Ball-balancing robots (ballbots) must lean to accelerate, and the lean is limited by the wheel torques and by safety. Planners that use the full robot model are expensive, while planners that ignore the balance dynamics produce references that cannot be followed. This paper presents a two-layer navigation scheme for a three-omni-wheel ballbot. The upper layer is a model predictive controller on a \emph{jerk model} of the ball position, with discrete-time control barrier functions for static, moving, rotating and concave obstacles. The lower layer is a hierarchical sliding mode controller with an extended state observer and a robust exponential control barrier function on the tilt (HSMC-ESO-ECBF). The layers are linked by an acceleration bound derived in closed form from the largest steady tilt that the barrier function admits, so that the plan never asks for a lean the tracker refuses. For obstacles whose motion is only predicted, the obstacles are inflated by conformal prediction radii that bound the prediction error with a prescribed probability. In MuJoCo, with omni-wheel roller contacts and torque saturation, the scheme keeps at least $10$\,cm of clearance, tracks the plan within $2$\,cm RMS, keeps the tilt below the limit and plans in $9$--$12$\,ms, $10$--$15$ times faster than a full-model nonlinear MPC that enters the inflated obstacles; with predicted obstacles, the conformal inflation removes the collisions of a constant-velocity predictor in $30$ random crossing scenarios and reduces them in an endless run on a closed path.\ifieee{} The source code and video illustrations are available at \url{https://safe-ballbot.github.io}.\fi

\end{abstract}



\section{Introduction}
\begin{figure}[!t]
\centering
\includegraphics[width=\hookwidth]{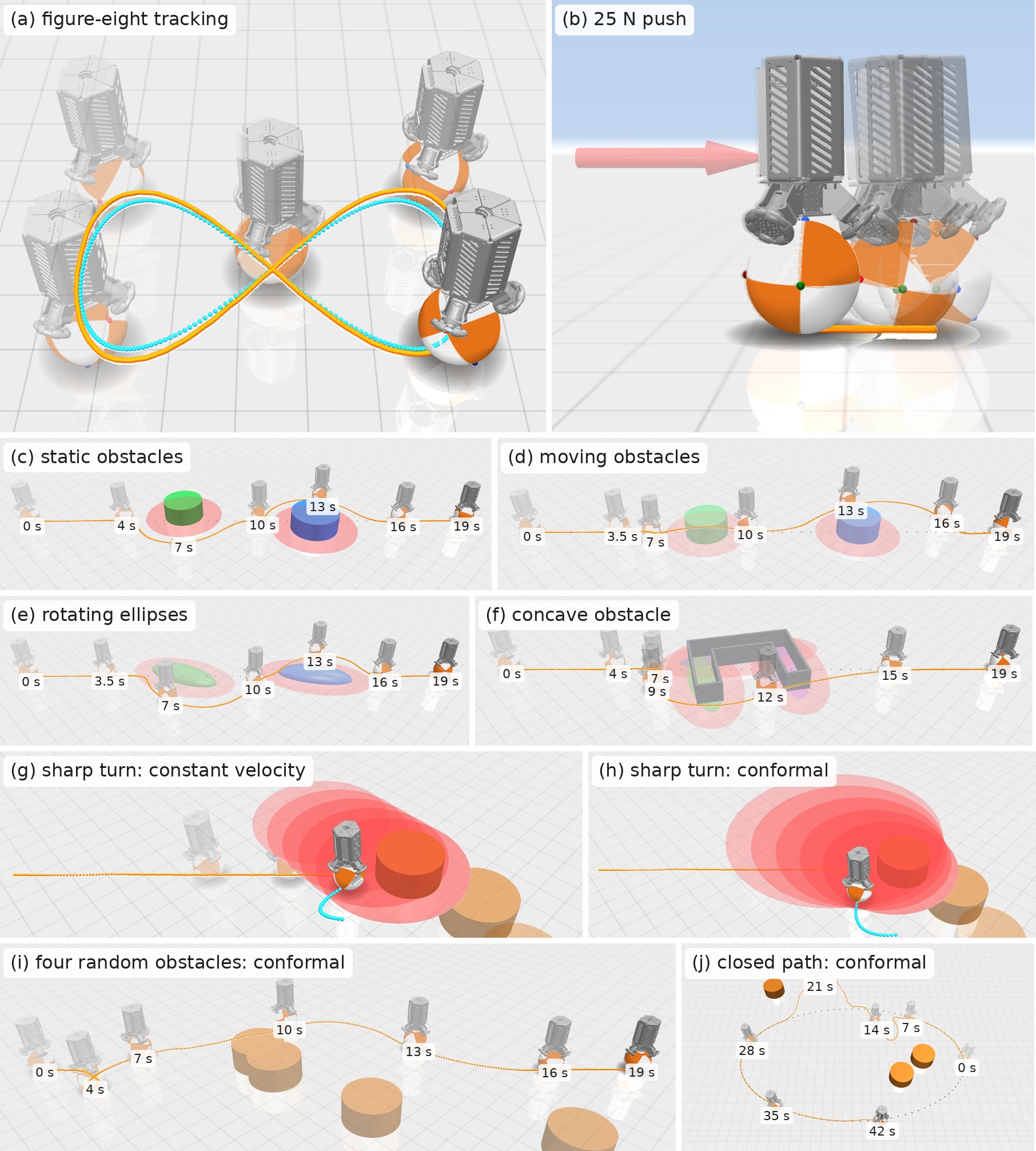}
\caption{The proposed scheme in MuJoCo. Tracking: (a) figure eight, (b) recovery from a $25$\,N push. Known obstacles: (c) static, (d) moving, (e) rotating, (f) concave. Predicted obstacles: an obstacle that turns sharply toward the path, with (g) constant-velocity and (h) conformal prediction, (i) four random obstacles crossing the path, (j) an endless run on a closed path among three roaming obstacles.}
\label{fig:hook}
\end{figure}

\begin{table*}[!t]
\centering
\begin{threeparttable}
\caption{Comparison with related work on ballbot motion planning and safe control.}
\label{tab:related}
\footnotesize\setlength{\tabcolsep}{3.6pt}\renewcommand{\arraystretch}{1.15}
\begin{tabular*}{\linewidth}{@{\extracolsep{\fill}}lllccccll@{}}
\toprule
& & & & \multicolumn{3}{c}{\hd{Obstacles}} & & \\
\cmidrule(lr){5-7}
\hd{Work} & \hd{Year} & \hd{Planner / controller model} & \hd{Online} & \hd{Static} & \hd{Moving} & \hd{Concave} & \hd{Tilt limit} & \hd{Evaluation} \\
\midrule
Nagarajan~\cite{nagarajan2013} & 2013 & shape-space planner & \no & \nr & \no & \no & \nr & hardware \\
Jespersen~\cite{jespersen2020} & 2020 & path-following MPC, reduced quaternion model & \yes & \yes & \no & \no & \yes{} soft & sim.\ + hardware \\
Mavrogiannis~\cite{mavrogiannis2021} & 2021 & topology-informed MPC & \yes & \nr & \yes{} crowd & \no & \nr & hardware + crowd sim. \\
Fischer~\cite{fischer2024} & 2024 & identified LTI model, MPC + LQR & \yes & \no & \no & \no & \nr & hardware \\
Vu~\cite{vu2024time} & 2024 & flat output, offline time-optimal + RRT*, ESO & \no & \yes & \no & \no & \yes{} ECBF & sim.\ (design model) \\
Pham~\cite{pham2025cbf} & 2025 & nonlinear MPC on the full 3D model, ESO & \yes & \yes & \yes & \yes & \yes{} CBF-QP & sim.\ (design model) \\
Salehi~\cite{salehi2025} & 2025 & model-free RL policy & \yes & \no & \no & \no & \no & MuJoCo \\
Wan~\cite{wan2026asroballet} & 2026 & friction-aware RL policy (sim-to-real) & \yes & \no & \no & \no & \no & MuJoCo + hardware \\
\midrule
\rowcolor{rowhl}\textbf{This work} & -- & \textbf{jerk model (robot-independent) + DCBF, ESO} & \yes & \yes & \yes & \yes & \yes{} ECBF, $\varphi^\star$ & MuJoCo, roller contacts \\
\bottomrule
\end{tabular*}
\begin{tablenotes}[flushleft]
\footnotesize
\item[] \yes: addressed; \no: not addressed; \nr: not reported. Online: replanning during the motion. ESO: extended state observer.
\end{tablenotes}
\end{threeparttable}
\end{table*}

\label{sec:intro}

\firstword{A}{ballbot} balances on a single spherical wheel driven by omni wheels. The design gives omnidirectional motion and a small footprint, but the system is underactuated: three wheel torques control five degrees of freedom (the ball position and the three body angles), and the ball can only be accelerated by first leaning the body in the direction of motion. Every motion planner for a ballbot therefore has to respect the balance dynamics, and every controller has to keep the body tilt in a safe range while it tracks the plan.

Recent works address these questions in two ways. Full-model approaches put the nonlinear robot dynamics into an online optimization, for example a nonlinear MPC with CBF constraints for moving obstacles~\cite{pham2025cbf} or a path-following MPC on a reduced quaternion model~\cite{jespersen2020}; they predict the balance dynamics but are computationally heavy and tie the planner to one robot model. Decoupled approaches plan offline, for example time-optimal trajectories in the flat-output domain tracked by an observer-based sliding mode controller with a tilt barrier function~\cite{vu2024time}; they are cheap to execute but cannot react to moving obstacles. Learning-based navigation~\cite{salehi2025} does not give explicit tilt or clearance guarantees.

This report studies a lighter split. The motion planner works on a \emph{jerk model} of the ball position, a chain of three integrators per axis that is independent of the robot dynamics. The only robot-specific information in the planner is the relation between acceleration and lean, which becomes an acceleration bound, and the robot footprint. Balancing, disturbance rejection and the hard tilt limit are left to the HSMC-ESO-ECBF tracker~\cite{vu2024time}. A key observation is that the robust barrier function of the tracker admits a steady tilt $\varphi^\star$ that is smaller than the tilt limit; the acceleration bound of the planner must be derived from $\varphi^\star$, not from the limit itself. All results are obtained on a MuJoCo model adapted from the asRoBallet platform~\cite{wan2026asroballet}, using its ballbot base with omni-wheel roller contacts and wheel torque saturation (\cref{fig:hook}).

The contributions are:
\begin{enumerate}
  \item \emph{Tracker} (\cref{sec:prelim}): an HSMC-ESO-ECBF controller for a three-omni-wheel ballbot with a sliding surface designed in closed form by zero placement, an ECBF on the tilt cone solved in closed form, and a state-dependent robust term, motivated by a measured input-gain mismatch of $30\,\%$ between the design model and the contact dynamics; the largest steady tilt that this robust ECBF admits is derived in closed form (\cref{prop:steady});
  \item \emph{Planner} (\cref{sec:main}): a jerk-model MPC with discrete-time CBFs for static, moving and rotating elliptic obstacles and for concave obstacles decomposed into convex pieces bounded by minimum-area ellipses, with a tracking margin for the low-level controller;
  \item \emph{Planner--tracker interface} (\cref{sec:main}): an acceleration bound derived from the steady tilt of the tracker \eqref{eq:amax}, so that the plan never asks for a lean the tracker refuses; it reduces the tracking error by a factor of $1.8$--$7$ compared to a bound derived from the tilt limit;
  \item \emph{Predicted obstacles} (\cref{sec:conformal}): an extension to obstacles whose motion is predicted rather than known, in which the obstacle ellipses are inflated by conformal prediction radii that bound the prediction error with a prescribed probability, and soft barrier constraints keep the planner feasible;
  \item \emph{Evaluation} (\cref{sec:sim}): a cross-validation of the planar, 3D (Kane) and MuJoCo models, and a simulation study of the complete scheme with static, moving, rotating, concave and predicted obstacles, including a comparison with a full-model NMPC, Monte Carlo runs over random scenarios and model mismatch.
\end{enumerate}

\cref{sec:related} reviews related work. \cref{sec:prelim} presents the robot, its kinematics and models, and the tracker. \cref{sec:main} develops the planner, and \cref{sec:conformal} extends it to predicted obstacles with conformal prediction. \cref{sec:sim} reports the simulation results, and \cref{sec:conclusion} concludes.

\section{Related Work}
\label{sec:related}

\ifieee\cref{tab:related} summarizes the planning and safety features of the ballbot works discussed below.

\fi\paragraph{Ballbot design and control} The first ballbot used an inverse mouse-ball drive and a two-loop PI--LQR controller~\cite{lauwers2006}; a decade of work on this platform is summarized in~\cite{nagarajan2014}. The three-omni-wheel design with a $45^\circ$ zenith angle drives the ball directly and adds control of the yaw axis~\cite{kumagai2009,fankhauser2010}. Linear designs, such as LQR with feedforward~\cite{fankhauser2010} or LQR inner loops with an MPC outer loop on an identified model~\cite{fischer2024}, are simple and run on hardware, but they lose accuracy away from the upright posture and treat the two planes as independent. Hierarchical sliding mode control (HSMC) on decoupled planes~\cite{pham2018} is robust to matched disturbances. Vu et al.~\cite{vu2024time} estimate the coupling between the planes and the model errors with an extended state observer (ESO)~\cite{han2009,guo2011}, in the spirit of active disturbance rejection control, and add an exponential CBF (ECBF)~\cite{nguyen2016,ames2017} that keeps the tilt inside a cone. Barrier Lyapunov functions~\cite{tee2009} also limit the tilt, but they are undefined outside the safe set and give no safe action once the limit is exceeded, for example after a push. None of these controllers plans around obstacles; they follow a given reference.

\paragraph{Motion planning for balancing robots} For a balancing robot, a planned path is feasible only if the body can lean as the path requires. Shape-space planning~\cite{nagarajan2013} maps desired accelerations to body shapes of the ballbot, and flatness-based planning~\cite{fliess1995,shomin2013,vu2024time} turns the planar ballbot into a chain of integrators, so that balance-consistent trajectories follow from smooth position curves; both are offline and do not handle moving obstacles. Online, Jespersen et al.~\cite{jespersen2020} run a path-following MPC on a reduced quaternion model with soft constraints for circular obstacles and for the tilt, validated on hardware. Mavrogiannis et al.~\cite{mavrogiannis2021} use an MPC that reasons about the topology of the robot's path relative to the people in a crowd. Pham et al.~\cite{pham2025cbf} use a nonlinear MPC on the full 3D model with discrete CBF constraints for moving and complex-shaped obstacles and a separate ESO-based CBF for the tilt; the full model makes each solve expensive, and since the tilt constraint is removed from the MPC, the plan may ask for leans that the tilt filter later rejects. Model-free reinforcement learning has been used for ballbot navigation on uneven terrain in MuJoCo~\cite{salehi2025}, without explicit safety constraints. Wan et al.~\cite{wan2026asroballet} close the sim-to-real gap of the humanoid ballbot asRoBallet with friction-aware reinforcement learning and deploy an end-to-end policy for velocity tracking and station keeping on hardware; obstacles and an explicit tilt limit are not considered. The MuJoCo model of their platform is the simulation model of the present work.

\paragraph{Reduced-order models for planning} Planning on a simple model and leaving the full dynamics to a tracking controller is common for other dynamic robots. For humanoids, the center-of-mass dynamics are approximated by a linear inverted pendulum and planned by preview control~\cite{kajita2003} or by a linear MPC with the center-of-mass jerk as input~\cite{wieber2006}; the jerk input keeps the references smooth and the problem a small quadratic program. For quadrotors, differential flatness turns the dynamics into chains of integrators, and minimum-snap~\cite{mellinger2011,richter2016} or minimum-jerk motion primitives~\cite{mueller2015} are planned on them and tracked by an attitude controller; the input limits of the full robot become bounds on the derivatives of the position. The ballbot has the same structure: by the lean--acceleration relation, the tilt limit becomes an acceleration bound on a triple integrator. What these works do not address is how to choose this bound when the tracker itself restricts the reachable tilt, as an ECBF does.

\paragraph{Control barrier functions and MPC} CBFs~\cite{ames2017} enforce forward invariance of a safe set through a constraint that is affine in the input, which leads to safety filters solved as quadratic programs; exponential CBFs~\cite{nguyen2016} extend this to constraints of higher relative degree such as a tilt limit. In MPC, discrete-time CBFs~\cite{zeng2021} replace distance constraints by a bound on the decay rate of the barrier along the horizon, so that the plan approaches an obstacle gradually and short horizons suffice. Polytopic obstacles can be handled exactly through duality-based distance constraints~\cite{thirugnanam2022}, at the price of extra variables per obstacle; bounding each convex piece by an ellipse~\cite{pham2025cbf} keeps one smooth constraint per piece.

\paragraph{Robust and observer-based barrier functions} A CBF guarantees safety only for the model it is designed on; model errors and disturbances enter the barrier condition directly. Robust CBFs~\cite{jankovic2018} add a margin that covers a bounded disturbance, and input-to-state safety~\cite{kolathaya2019} quantifies how far the state may leave the safe set when the disturbance is not covered. When the disturbance is estimated instead, an observer such as the ESO provides the drift term of the barrier condition, and only the residual estimation error needs a margin~\cite{vu2024time,pham2025cbf}. In the ballbot tracker, this residual is covered by the state-dependent term $2\delta\|\vect\varphi\|_1$, which vanishes at the upright posture. These works study the safety of the filtered system; the side effect of the margin on the set of postures the filter accepts in steady state, and therefore on the motions that a planner may request, has not been analyzed. \cref{prop:steady} gives this set in closed form.

\paragraph{Planning among predicted obstacles} When the future motion of an obstacle is not known, the planner must use a prediction and account for its error. Chance-constrained MPC bounds the collision probability under an assumed, typically Gaussian, distribution of the obstacle position~\cite{zhu2019chance}; the guarantee then depends on how well this distribution matches the real motion. Simple predictors remain competitive: a constant-velocity model is a strong baseline for pedestrian motion~\cite{scholler2020constant}. Conformal prediction~\cite{vovk2005,angelopoulos2021} instead bounds the error of any predictor with a prescribed probability, without a model of its distribution, from a calibration set of past prediction errors. Lindemann et al.~\cite{lindemann2023conformal} use these bounds as distance margins in an MPC for planning among dynamic agents, and adaptive variants recalibrate them online under distribution shift~\cite{gibbs2021adaptive,dixit2023adaptive}. These works plan for point-mass or simple vehicle models; the present work inflates the barrier functions of a ballbot planner with the conformal radii and keeps the tilt-consistent interface to the tracker unchanged.

\paragraph{Layered planning with tracking-error bounds} When a planner on a simple model is tracked by a controller of the full robot, safety requires a bound on the tracking error. Tube MPC~\cite{mayne2005} tightens the constraints of the nominal plan by a robust invariant set of the error, and FaSTrack~\cite{herbert2017} computes a tracking-error bound by reachability analysis and inflates the obstacles by it. The present scheme follows this structure with a simpler, empirical margin $r_{\text{track}}$, and adds a second interface that these frameworks do not have: the planner's acceleration bound is derived from the tilt that the tracker's barrier function actually admits.

\section{Preliminaries}
\label{sec:prelim}

\subsection{Robot and assumptions}
\label{sec:robot}

\begin{figure}[!t]
\centering
\ifieee\includegraphics[width=\figw]{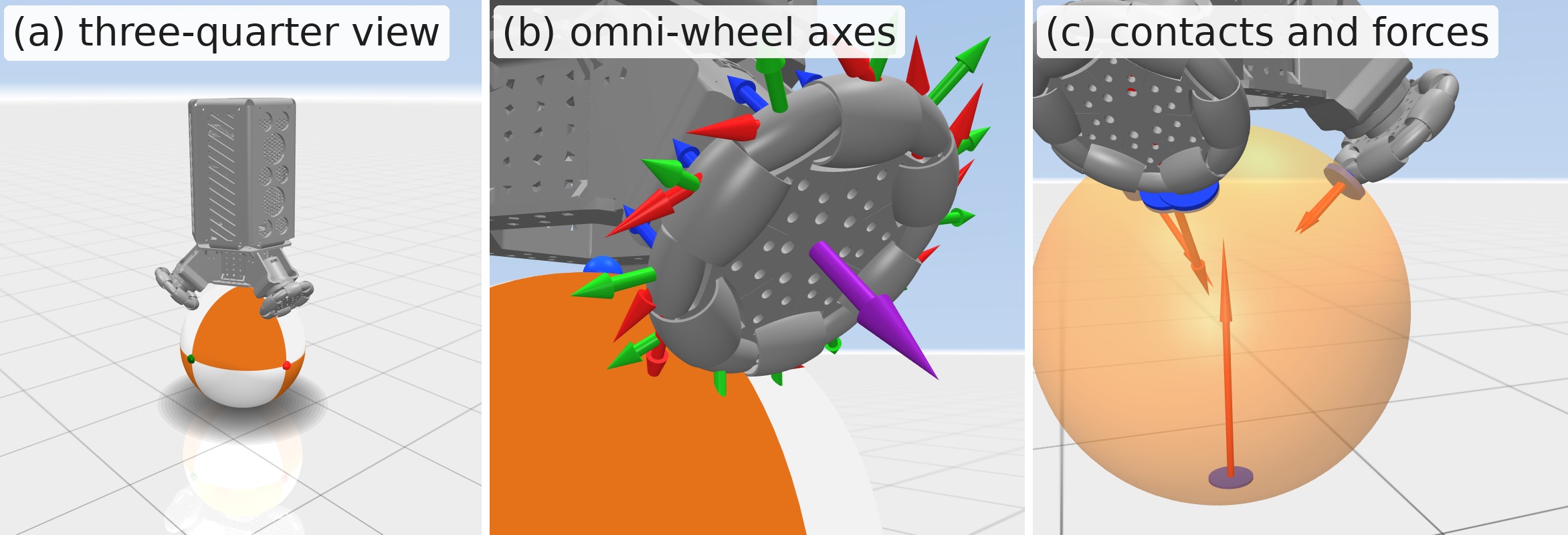}\fi
\caption{\ifieee The ballbot in MuJoCo: (a) balancing robot, (b) omni wheel~1 with its axis $\vect a_1$ (violet) and the roller frames, (c) contact points (blue) and forces (red).\fi}
\label{fig:sim_overview}
\end{figure}

The robot model is adapted from the asRoBallet ballbot of Wan et al.~\cite{wan2026asroballet}: a body that balances on a ball of radius $R$ driven by three omni wheels evenly spaced around the vertical axis, with inclined wheel axles (\cref{fig:sim_overview}); the humanoid upper body is removed. We use the authors' released MJCF model from the asRoBallet MuJoCo repository\footnote{\url{https://github.com/asRoBallet/asRoBallet_mujoco}} and retain its ball, omni-wheel and roller-contact geometry. The base has a free joint and the ball is connected to it by a ball joint. Each omni wheel has one actuated axle joint with a bounded torque and a set of passive rollers.

\begin{assumption}[Design models]
\label{as:design}
The design models rest on the following simplifications:
\begin{enumerate}
\renewcommand{\theenumi}{A\arabic{enumi}}\renewcommand{\labelenumi}{(\theenumi)}
  \item\label{as:roll} the ball rolls on the floor without slipping, and the omni wheels transmit torque to the ball ideally;
  \item\label{as:rigid} the base and the wheels form a single rigid body;
  \item\label{as:planes} near the upright posture the motion splits into the $x$--$z$ plane, the $y$--$z$ plane and the yaw axis (the 3D model of \cref{sec:model3d} drops this assumption);
  \item\label{as:friction} friction, wheel and roller inertia, and roller slip are neglected.
\end{enumerate}
\end{assumption}
The MuJoCo model used for all results contains none of these simplifications.

\subsection{Omni-wheel kinematics}
\label{sec:kinematics}

\paragraph{Wheel geometry} In the base frame, wheel $i \in \{1, 2, 3\}$ touches the ball at azimuth $\alpha_i = 2\pi(i-1)/3$ and zenith angle $\beta$, measured from the upward vertical at the ball center. The unit vector from the ball center to the contact point and the unit rotation axis of the wheel are
\begin{equation}
  \hat{\vect c}_i = \begin{bmatrix}\sin\beta\cos\alpha_i\\ \sin\beta\sin\alpha_i\\ \cos\beta\end{bmatrix},\qquad
  \vect a_i = \begin{bmatrix}\cos\beta\cos\alpha_i\\ \cos\beta\sin\alpha_i\\ -\sin\beta\end{bmatrix},
  \label{eq:wheel_geometry}
\end{equation}
so that $\vect a_i \perp \hat{\vect c}_i$: the axis lies in the vertical plane through the contact point, tilted by $\beta$ from the horizontal. At the contact, the wheel rim moves along $\vect t_i = \vect a_i \times \hat{\vect c}_i$ (a horizontal direction tangent to the circle of contacts); the passive rollers leave the ball free to slide along $\vect a_i$.


\ifieee
\paragraph{Velocity and torque maps} With $\vect\omega$ the angular velocity of the ball relative to the body (base frame), $r_w$ the wheel radius and $\dot\phi_i$ the wheel speeds, no slip along the rim direction $\vect t_i$ gives
\begin{equation}
  \dot{\vect\phi} = \mat J\,\vect\omega,\qquad
  \mat J = -\frac{R}{r_w}
  \begin{bmatrix}\vect a_1^\top\\ \vect a_2^\top\\ \vect a_3^\top\end{bmatrix},
\label{eq:J}
\end{equation}
while the rollers absorb the contact velocity along $\vect a_i$. For evenly spaced wheels the singular values of $\mat J$ are $(R/r_w)\sqrt{3/2}\cos\beta$ (twice) and $(R/r_w)\sqrt3\sin\beta$, so $\mat J$ is invertible for every $0 < \beta < \pi/2$. By the balance of power, $\vect\tau_w^\top\dot{\vect\phi} = \vect\tau_{\text{ball}}^\top\vect\omega$, the torque that the body applies to the ball and the wheel torques are related by
\begin{equation}
  \vect\tau_{\text{ball}} = \mat J^\top\vect\tau_w
  \quad\Longleftrightarrow\quad
  \vect\tau_w = \mat J^{-\top}\vect\tau_{\text{ball}} ;
  \label{eq:torque_map}
\end{equation}
the controllers compute $\vect\tau_{\text{ball}}$ in the world frame and rotate it into the base frame before \eqref{eq:torque_map} is applied.
\fi

\subsection{Planar model}
\label{sec:planar}

With $\vect q = [\psi,\ \varphi]^\top$ (ball rolling angle, $x = R\psi$, and body tilt) and the torque $\tau$ applied by the body to the ball, the Lagrange equations are
\begin{equation}
\begin{gathered}
  \mat M(\varphi)
  \begin{bmatrix}\ddot\psi\\ \ddot\varphi\end{bmatrix}
  =
  \begin{bmatrix}
    \tau + MRl\sin\varphi\,\dot\varphi^2\\
    -\tau + Mgl\sin\varphi
  \end{bmatrix},\\
  \mat M(\varphi) = \begin{bmatrix}
    (m_b+M)R^2 + I_b & MRl\cos\varphi\\
    MRl\cos\varphi & Ml^2 + I_B
  \end{bmatrix}.
\end{gathered}
\label{eq:nonlinear}
\end{equation}
The $y$--$z$ plane has the same equations with $y = -R\psi$, $\varphi$ = roll and $I_{B,x}$, and the yaw axis is $I_{B,z}\ddot\gamma = -\tau_z$.

Linearized about $\varphi = 0$ with $\vect s = [\psi,\ \varphi,\ \dot\psi,\ \dot\varphi]^\top$,
\begin{equation}
\begin{gathered}
  \dot{\vect s} = \mat A\vect s + \mat B\tau,\qquad
  \mat A = \begin{bmatrix}
    0 & 0 & 1 & 0\\
    0 & 0 & 0 & 1\\
    0 & a_1 & 0 & 0\\
    0 & a_2 & 0 & 0
  \end{bmatrix},\quad
  \mat B = \begin{bmatrix}0\\0\\b_1\\b_2\end{bmatrix},\\
  \begin{bmatrix}a_1\\a_2\end{bmatrix} = \mat M(0)^{-1}\begin{bmatrix}0\\Mgl\end{bmatrix},\qquad
  \begin{bmatrix}b_1\\b_2\end{bmatrix} = \mat M(0)^{-1}\begin{bmatrix}1\\-1\end{bmatrix},
\end{gathered}
\label{eq:linear}
\end{equation}
with eigenvalues $\{0, 0, \pm\sqrt{a_2}\}$; since $a_2 > 0$ the body is an unstable inverted pendulum.

\paragraph{Lean and acceleration} For a constant ball acceleration $\ddot\psi$ with $\ddot\varphi = 0$, the linear model gives the equilibrium lean and torque
\begin{equation}
\begin{gathered}
  \varphi = \frac{M_{11} + M_{21}}{Mgl}\,\ddot\psi = \kappa\,\ddot x,\qquad
  \tau = M_{11}\ddot\psi,\\
  \kappa = \frac{M_{11} + M_{21}}{R\,Mgl} ,
\end{gathered}
\label{eq:lean}
\end{equation}
where $M_{ij}$ are the entries of $\mat M(0)$. This relation is the feedforward of the trackers and the basis of the acceleration bound of the planner.

\subsection{Nonlinear 3D model}
\label{sec:model3d}

To check the decoupled-plane assumption, a 3D rigid-body model keeps the coupling between the planes and yaw, the gyroscopic terms, the products of inertia and the COM offset $\vect r_B$. The body pivots about the ball center; the ball rolls without slipping and does not spin about the vertical.

\ifieee
With the ball position $(x, y)$ and the unit quaternion $\vect\zeta$ of the body as coordinates and $\vect u = [v_x,\ v_y,\ \vect\omega^\top]^\top \in \mathbb R^5$ (ball velocity and body angular velocity) as generalized speeds, projecting the Newton--Euler equations of the ball and the body on the partial velocities (Kane's method) gives
\begin{equation}
  \mat{\mathcal M}(\vect\zeta)\,\dot{\vect u} + \mat{\mathcal C}(\vect\zeta, \vect u)\,\vect u
  + \mat{\mathcal G}(\vect\zeta) = \mat{\mathcal B}(\vect\zeta)\,\vect\tau_w ,
  \label{eq:kane}
\end{equation}
with the mass, Coriolis, gravity and input terms of the two rigid bodies; the input term maps the wheel torques through \eqref{eq:torque_map}. For a COM on the body axis, diagonal inertia and small tilts, the linearization of \eqref{eq:kane} reduces to \eqref{eq:linear} in each plane and to $I_{B,z}\ddot\gamma = -\tau_z$ for the yaw; the 3D model thus differs from the planar models only through the coupling terms, the COM offset and the products of inertia, which grow with the tilt.
\fi

\subsection{Tracking controller: HSMC-ESO-ECBF}
\label{sec:hsmc}

The tracker follows~\cite{vu2024time}: per plane an ESO and an HSMC, and one ECBF on the tilt cone for both planes; the yaw axis uses LQR.

\paragraph{Plane model with total disturbance} For each plane, with $\vect q = [\psi,\ \varphi]^\top$ and the virtual ball torque $u$ of that plane, \eqref{eq:nonlinear} is written as
\begin{equation}
  \ddot{\vect q} = \vect f + \vect g(\varphi)\,u,\qquad
  \vect g(\varphi) = \mat M(\varphi)^{-1}\begin{bmatrix}1\\-1\end{bmatrix},
  \label{eq:plane_dist}
\end{equation}
where the total disturbance $\vect f$ collects gravity, centrifugal terms, the coupling with the other plane and with yaw, the omni-wheel contact and all model errors; $\vect g(0) = [b_1,\ b_2]^\top$ of \eqref{eq:linear}. Lumping everything but the input channel into one unknown signal is the idea of active disturbance rejection control~\cite{han2009}; only $\vect g$ has to be known.

\paragraph{Extended state observer} The total disturbance is estimated as an extra state by a linear high-gain ESO~\cite{han2009,guo2011,vu2024time}, from the measured $\vect q$ only:
\begin{equation}
\begin{aligned}
  \dot{\hat{\vect q}}_1 &= \hat{\vect q}_2 + \tfrac{\gamma_1}{\varepsilon}\vect e,\\
  \dot{\hat{\vect q}}_2 &= \hat{\vect q}_3 + \tfrac{\gamma_2}{\varepsilon^2}\vect e + \vect g(\hat\varphi)\,u,\\
  \dot{\hat{\vect q}}_3 &= \tfrac{\gamma_3}{\varepsilon^3}\vect e,\qquad
  \vect e = \vect q - \hat{\vect q}_1 ,
\end{aligned}
\label{eq:eso}
\end{equation}
where $s^3 + \gamma_1 s^2 + \gamma_2 s + \gamma_3$ is Hurwitz and $\varepsilon > 0$ sets the observer bandwidth (the poles scale with $1/\varepsilon$). If $\dot{\vect f}$ is bounded, the estimation errors are ultimately bounded and shrink with $\varepsilon$~\cite{guo2011}; a small $\varepsilon$ gives a fast estimate but amplifies measurement noise. The input fed to the ESO is the ball torque that the wheels actually produce \emph{after} saturation; otherwise $\hat{\vect q}_3$ absorbs the saturation error and winds up.

\paragraph{Hierarchical sliding mode control} The ballbot plane has one input for two outputs, the ball position and the tilt. Hierarchical sliding mode control~\cite{pham2018,vu2024time} builds one sliding surface per output and combines them into a single surface for the input. With $e_1 = \hat\psi - \psi_r$, $e_2 = \hat\varphi - \varphi_r$, $s_1 = \dot e_1 + k_1 e_1$, $s_2 = \dot e_2 + k_2 e_2$, $s = k_\alpha s_1 + k_\beta s_2$ and the exponential reaching law~\cite{gao1993} $\dot s = -k s - \eta\,\mathrm{sat}(s/\Phi)$, whose boundary layer $\Phi$ avoids chattering,
\begin{equation}
\begin{gathered}
  u_{\text{HSMC}} = -\frac{\Xi + k s + \eta\,\mathrm{sat}(s/\Phi)}{k_\alpha g_1 + k_\beta g_2},\\
  \Xi = k_\alpha(\hat f_1 - \ddot\psi_r + k_1\dot e_1) + k_\beta(\hat f_2 - \ddot\varphi_r + k_2\dot e_2),
\end{gathered}
\label{eq:hsmc}
\end{equation}
where the tilt reference is the lean \eqref{eq:lean} of the reference acceleration.

\begin{proposition}[Sliding surface by zero placement]
\label{prop:surface}
Write $s = \vect c^\top[e_1,\ e_2,\ \dot e_1,\ \dot e_2]^\top$ with $\vect c = [k_\alpha k_1,\ k_\beta k_2,\ k_\alpha,\ k_\beta]^\top$, and let the linearized plane be \eqref{eq:linear} with $\mat A_{21} = \bigl[\begin{smallmatrix}0 & a_1\\
0 & a_2\end{smallmatrix}\bigr]$ and $\mat B = [0,\ 0,\ b_1,\ b_2]^\top$. The motion on $s = 0$ has the poles $-p_1, -p_2, -p_3$ if
\begin{equation}
\begin{aligned}
  c_3 &= \frac{\sigma_2\,\beta}{\kappa_0}, &
  c_1 &= \frac{\sigma_3\,\beta}{\kappa_0}, &
  \kappa_0 &= a_1 b_2 - b_1 a_2,\\
  c_4 &= \frac{\beta - c_3 b_1}{b_2}, &
  c_2 &= \frac{\sigma_1\,\beta - c_1 b_1}{b_2}, &&
\end{aligned}
\end{equation}
where $s^3 + \sigma_1 s^2 + \sigma_2 s + \sigma_3 = (s+p_1)(s+p_2)(s+p_3)$ and $\beta = \vect c^\top \mat B \neq 0$.
\end{proposition}
\ifieee
\begin{proof}
With $\vect x = [e_1,\ e_2,\ \dot e_1,\ \dot e_2]^\top$, the motion on $s = \vect c^\top\vect x = 0$ has as poles the zeros of $H(\lambda) = \vect c^\top(\lambda \mat I - \mat A)^{-1}\mat B$. A direct computation gives $H(\lambda) = n(\lambda)/(\lambda^2(\lambda^2 - a_2))$ with $n(\lambda) = \beta\lambda^3 + (c_1 b_1 + c_2 b_2)\lambda^2 + c_3\kappa_0\lambda + c_1\kappa_0$, and equating $n$ with $\beta(\lambda^3 + \sigma_1\lambda^2 + \sigma_2\lambda + \sigma_3)$ gives the stated gains.
\end{proof}
\fi
The sign of $\beta$ is chosen such that all four gains are positive, as required in~\cite{vu2024time}.

\paragraph{Exponential control barrier function} The safe set is the tilt cone $\mathcal C = \{h \ge 0\}$, $h = \theta_{\max}^2 - \varphi_x^2 - \varphi_y^2$. Since the input appears only in $\ddot h$ ($h$ has relative degree two), an exponential CBF~\cite{nguyen2016} is used, with a robust term in the spirit of~\cite{jankovic2018}:
\begin{equation}
  \ddot h + \ell_2\dot h + \ell_1 h \ge 2\delta\,\|\vect\varphi\|_1,
  \label{eq:ecbf}
\end{equation}
with $s^2 + \ell_2 s + \ell_1$ Hurwitz. If the predicted tilt acceleration $\hat{\vect f}_\varphi + \mat G_\varphi\vect u$ is wrong by at most $\delta$ per axis, the error enters $\ddot h$ as $-2\vect\varphi^\top\vect e \ge -2\delta\|\vect\varphi\|_1$, which the right-hand side covers; unlike a constant margin, the term vanishes at the upright posture. Condition \eqref{eq:ecbf} is affine in the two planar inputs, $\vect a^\top\vect u \ge b$, with $\vect a = -2\,\mat G_\varphi\vect\varphi$ and $b = 2\|\dot{\vect\varphi}\|^2 + 2\,\vect\varphi^\top\hat{\vect f}_\varphi - \ell_2\dot h - \ell_1 h + 2\delta\|\vect\varphi\|_1$, and the safety filter $\min_{\vect u}\|\vect u - \vect u_{\text{HSMC}}\|^2$ s.t.\ $\vect a^\top\vect u \ge b$ has the closed-form solution
\begin{equation}
  \vect u = \vect u_{\text{HSMC}}
  + \frac{\max(0,\ b - \vect a^\top\vect u_{\text{HSMC}})}{\|\vect a\|^2}\,\vect a .
  \label{eq:ecbf_qp}
\end{equation}

\paragraph{Steady tilt admitted by the filter} The robust term protects the tilt limit, but it also changes which postures the filter accepts indefinitely, and this matters for the planner. A planned path with a constant acceleration $\vect a$ asks, by \eqref{eq:lean}, for a constant lean $\vect\varphi = \kappa\vect a$ held for as long as the acceleration lasts. Such a lean can be sustained only if \eqref{eq:ecbf} keeps holding while the body stays at it; otherwise the filter modifies the input, the body leans back, and the robot falls behind the plan. For a steady lean, $\dot{\vect\varphi} = \ddot{\vect\varphi} = \vect 0$, the derivatives of the barrier vanish,
\begin{equation}
  \dot h = -2\vect\varphi^\top\dot{\vect\varphi} = 0,\qquad
  \ddot h = -2\|\dot{\vect\varphi}\|^2 - 2\vect\varphi^\top\ddot{\vect\varphi} = 0,
\end{equation}
and \eqref{eq:ecbf} reduces to a condition on the lean alone,
\begin{equation}
  \ell_1\bigl(\theta_{\max}^2 - \|\vect\varphi\|^2\bigr) \ \ge\ 2\delta\,\|\vect\varphi\|_1 .
  \label{eq:steady_condition}
\end{equation}
Without the robust term ($\delta = 0$), \eqref{eq:steady_condition} is the tilt cone itself and every lean up to $\theta_{\max}$ is admitted. With $\delta > 0$ the right-hand side grows with the lean, so the admitted set shrinks; since $\|\vect\varphi\|_2 \le \|\vect\varphi\|_1 \le \sqrt2\,\|\vect\varphi\|_2$, it shrinks most for a lean along a diagonal between the two planes.

\begin{proposition}[Steady tilt admitted by the robust ECBF]
\label{prop:steady}
Let $\vect\varphi$ be a constant lean. (i) For a lean in one plane, $\vect\varphi = (\varphi, 0)$ or $(0, \varphi)$, the robust ECBF condition \eqref{eq:ecbf} holds if and only if
\begin{equation}
  |\varphi| \le \varphi^\star = -\frac{\delta}{\ell_1} + \sqrt{\frac{\delta^2}{\ell_1^2} + \theta_{\max}^2}.
  \label{eq:phistar}
\end{equation}
(ii) For a lean in any direction, \eqref{eq:ecbf} holds if
\begin{equation}
  \|\vect\varphi\| \le \bar\varphi^\star = -\frac{\sqrt2\,\delta}{\ell_1} + \sqrt{\frac{2\delta^2}{\ell_1^2} + \theta_{\max}^2},
  \label{eq:phistar_any}
\end{equation}
and the bound is attained along the diagonals. (iii) For $\delta > 0$, $\bar\varphi^\star < \varphi^\star < \theta_{\max}$.
\end{proposition}
\begin{proof}
For a steady lean, \eqref{eq:ecbf} is equivalent to \eqref{eq:steady_condition}, so it suffices to analyze this condition. In a single plane, let $z = |\varphi| \ge 0$. Then $\|\vect\varphi\|_1 = \|\vect\varphi\| = z$, and \eqref{eq:steady_condition} reads
\begin{equation*}
  p_\delta(z) = \ell_1 z^2 + 2\delta z - \ell_1\theta_{\max}^2 \ \le\ 0 .
\end{equation*}
The quadratic $p_\delta$ is convex, and $p_\delta(0) = -\ell_1\theta_{\max}^2 < 0$, so its two roots have opposite signs. On $z \ge 0$, $p_\delta(z) \le 0$ holds exactly up to the positive root, which is $\varphi^\star$ of \eqref{eq:phistar}.

For an arbitrary lean direction, let $z = \|\vect\varphi\|$. Since $\|\vect\varphi\|_1 \le \sqrt2\,z$,
\begin{equation*}
  \ell_1\bigl(\theta_{\max}^2 - z^2\bigr) - 2\delta\|\vect\varphi\|_1
  \ \ge\ -p_{\sqrt2\delta}(z),
\end{equation*}
so \eqref{eq:steady_condition} holds whenever $p_{\sqrt2\delta}(z) \le 0$, i.e.\ by the single-plane argument with $\delta$ replaced by $\sqrt2\,\delta$, whenever $z \le \bar\varphi^\star$. Along a diagonal, $\|\vect\varphi\|_1 = \sqrt2\,z$ and the inequality above is an equality, so the bound is attained.

To compare the two bounds, for $c > 0$, the positive root $\rho(c) = -c + \sqrt{c^2 + \theta_{\max}^2}$ satisfies
\begin{equation*}
  \rho'(c) = -1 + \frac{c}{\sqrt{c^2 + \theta_{\max}^2}} < 0,\qquad
  \rho(c) < \theta_{\max},
\end{equation*}
the latter because $\sqrt{c^2 + \theta_{\max}^2} < c + \theta_{\max}$. With $c = \delta/\ell_1$ and $c = \sqrt2\,\delta/\ell_1$ this gives $\bar\varphi^\star < \varphi^\star < \theta_{\max}$.
\end{proof}
A steady acceleration that needs more lean than $\varphi^\star$ (in one plane) or $\bar\varphi^\star$ (in the worst direction) cannot be sustained by the tracker, even though it is below $\theta_{\max}$. The gap grows with the robustness level $\delta/\ell_1$; it is the price of the robust term, and the planner must respect it (\cref{sec:main}).

\ifieee
\begin{remark}[Comparison with the tracker of Vu et al.~\cite{vu2024time}]
\label{rem:vu2024}
The structure of the tracker is that of~\cite{vu2024time}. The present design assigns the sliding dynamics in closed form (\cref{prop:surface}) instead of tuning the gains, replaces the switching term by a saturation to avoid chattering, covers the estimation error with the single robust term $2\delta\|\vect\varphi\|_1$ and a closed-form filter instead of a quadratic program with Lipschitz-based bounds (a weaker but practical guarantee), analyzes the steady tilt that this term admits (\cref{prop:steady}), feeds the ESO with the saturated torque and measures the tilt toward the body COM. Unlike~\cite{vu2024time}, which derives the tilt reference exactly from a flat-output trajectory, the planner here provides only the quasi-static lean \eqref{eq:lean}.
\end{remark}
\fi

\subsection{Baseline: LQR}
\label{sec:lqr}

The baseline is the common linear design for ballbots~\cite{fankhauser2010}: a discrete LQR per plane on \eqref{eq:linear} (zero-order hold at the control period), $u = \tau_{ff} - \mat K(\vect s - \vect s_r)$ with weights $\mat Q \succ 0$, $R_u > 0$. The reference lean and the feedforward torque follow \eqref{eq:lean}. The yaw axis uses the same design on $I_{B,z}\ddot\gamma = -\tau_z$.

\section{Jerk-Model MPC with Control Barrier Functions}
\label{sec:main}

\subsection{Architecture and design rationale}

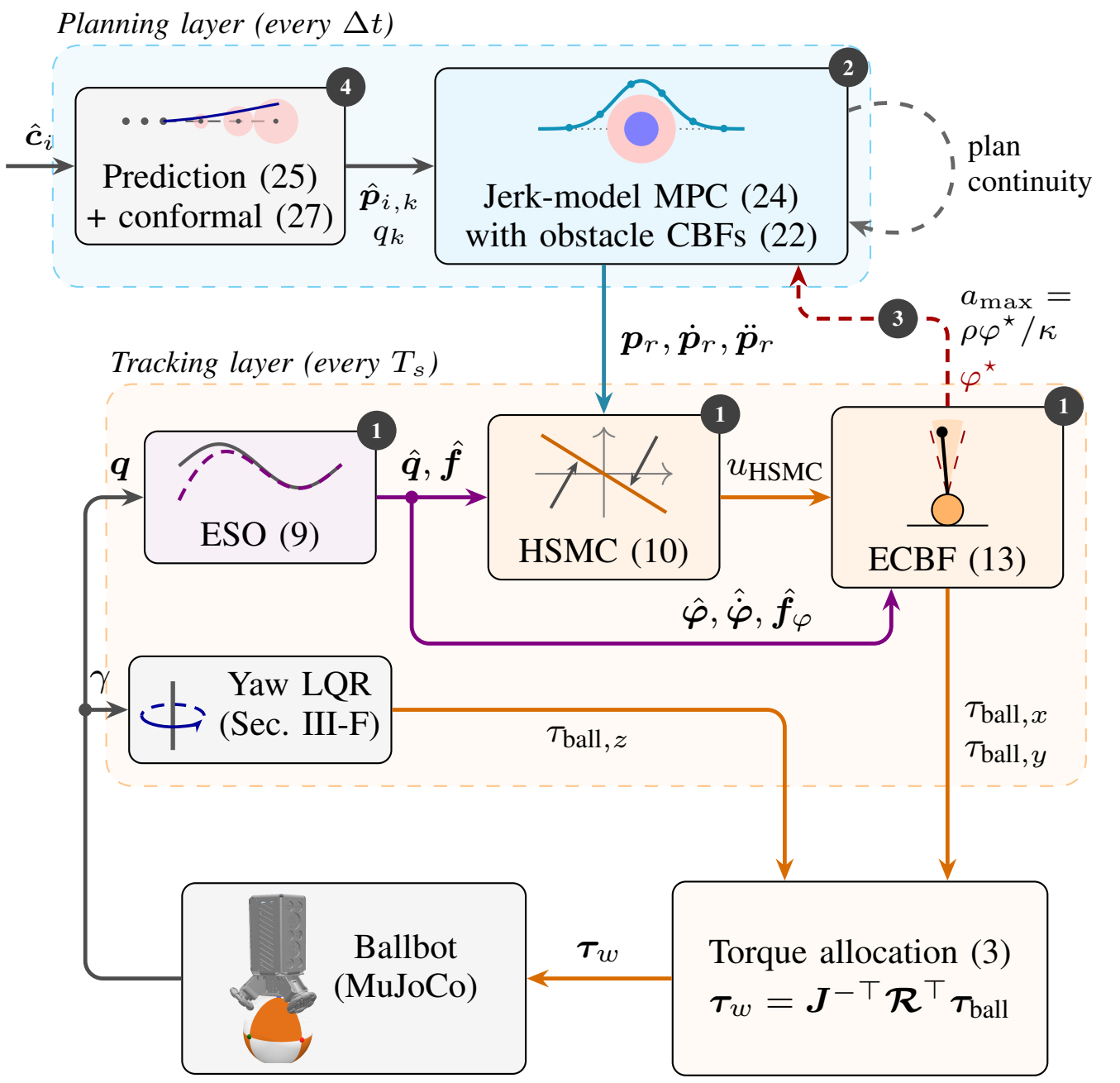
\begin{figure}[!t]
\centering
\newsavebox{\iconobs}\newsavebox{\iconmpc}\newsavebox{\iconeso}\newsavebox{\iconsmc}%
\newsavebox{\iconcbf}\newsavebox{\iconyaw}%
\sbox{\iconobs}{\begin{tikzpicture}[scale=0.5]
  \foreach \x in {0, 0.3, 0.6} \fill[gray!70!black] (\x, 0) circle (0.06);
  \draw[densely dashed, gray!70!black] (0.6, 0) -- (2.4, 0);
  \foreach \x/\q in {1.2/0.12, 1.8/0.24, 2.4/0.36} {
    \fill[red!25, fill opacity=0.6] (\x, 0) circle (\q);
    \fill[gray!70!black] (\x, 0) circle (0.04);
  }
  \draw[blue!60!black, line width=0.6pt] (0.6, 0) .. controls (1.4, 0.05) and (2.0, 0.2) .. (2.45, 0.28);
\end{tikzpicture}}%
\sbox{\iconmpc}{\begin{tikzpicture}[scale=0.55]
  \draw[densely dotted, gray] (0, 0) -- (3.0, 0);
  \fill[red!20] (1.5, 0) circle (0.5);
  \fill[blue!50] (1.5, 0) circle (0.25);
  \draw[cyan!70!black, thick, domain=0:3.0, samples=60] plot (\x, {0.7*exp(-((\x-1.5)/0.55)^2)});
  \foreach \x in {0.45, 0.9, 1.35, 1.8, 2.25, 2.7}
    \fill[cyan!70!black] (\x, {0.7*exp(-((\x-1.5)/0.55)^2)}) circle (0.055);
\end{tikzpicture}}%
\sbox{\iconeso}{\begin{tikzpicture}[scale=0.5]
  \draw[gray!70!black, thick, domain=0:2.6, samples=50] plot (\x, {0.35*sin(\x*140)});
  \draw[violet, thick, densely dashed, domain=0:2.6, samples=50]
    plot (\x, {0.35*sin(\x*140) - 0.55*exp(-2.2*\x)});
\end{tikzpicture}}%
\sbox{\iconsmc}{\begin{tikzpicture}[scale=0.5]
  \draw[->, gray, line width=0.4pt] (-1.1, 0) -- (1.1, 0);
  \draw[->, gray, line width=0.4pt] (0, -0.7) -- (0, 0.75);
  \draw[orange!80!black, thick] (-1.0, 0.62) -- (1.0, -0.62);
  \draw[-{Stealth[length=3pt, width=2.4pt]}, line width=0.7pt, black!70] (0.85, 0.6) -- (0.42, -0.2);
  \draw[-{Stealth[length=3pt, width=2.4pt]}, line width=0.7pt, black!70] (-0.85, -0.55) -- (-0.42, 0.2);
\end{tikzpicture}}%
\sbox{\iconcbf}{\begin{tikzpicture}[scale=0.5]
  \fill[orange!25] (0, 0) -- (102:1.2) arc (102:78:1.2) -- cycle;
  \draw[dashed, red!60!black] (0, 0) -- (105:1.25) (0, 0) -- (75:1.25);
  \draw[thick] (0, 0) -- (95:1.0);
  \fill (95:1.0) circle (0.09);
  \draw[fill=orange!60] (0, -0.25) circle (0.25);
  \draw (-0.65, -0.5) -- (0.65, -0.5);
\end{tikzpicture}}%
\sbox{\iconyaw}{\begin{tikzpicture}[scale=0.5]
  \draw[densely dashed, blue!60!black, line width=0.7pt] (0.5, 0.4) arc (0:180:0.5 and 0.18);
  \draw[line width=1pt, gray!70!black] (0, -0.15) -- (0, 0.95);
  \draw[-{Stealth[length=3pt, width=2.4pt]}, blue!60!black, line width=0.7pt] (-0.5, 0.4) arc (180:340:0.5 and 0.18);
\end{tikzpicture}}%
\begin{tikzpicture}[>={Stealth[length=5pt, width=4pt]}, font=\footnotesize,
  box/.style={draw, rounded corners=3pt, align=center, minimum width=#1, minimum height=0.9cm, inner sep=2.5pt},
  box/.default=1.85cm,
  sig/.style={line width=0.9pt, rounded corners=4pt},
  ref/.style={sig, cyan!60!black},
  trq/.style={sig, orange!85!black},
  meas/.style={sig, gray!60!black},
  est/.style={sig, violet},
  design/.style={sig, red!65!black, densely dashed},
  badge/.style={circle, fill=black!75, text=white, font=\tiny\bfseries, inner sep=0.9pt, minimum size=9pt},
  layer/.style={draw=#1!45, fill=#1!4, rounded corners=6pt, dashed, inner sep=5pt}]
  \node[box=1.85cm, fill=gray!8]  (obs) at (-3.15, 0)   {\usebox{\iconobs}\\[1pt]Prediction \eqref{eq:cv_predictor}\\+ conformal \eqref{eq:conformal_radius}};
  \node[box=3.3cm, fill=cyan!8]   (mpc) at (0.3, 0)     {\usebox{\iconmpc}\\[1pt]Jerk-model MPC \eqref{eq:mpc}\\with obstacle CBFs \eqref{eq:dcbf}};
  \node[box, fill=violet!6]       (eso)  at (-2.75, -2.65) {\usebox{\iconeso}\\[1pt]ESO \eqref{eq:eso}};
  \node[box, fill=orange!10]      (smc)  at (0, -2.65)     {\usebox{\iconsmc}\\[1pt]HSMC \eqref{eq:hsmc}};
  \node[box, fill=orange!10]      (cbf)  at (2.75, -2.65)  {\usebox{\iconcbf}\\[1pt]ECBF \eqref{eq:ecbf_qp}};
  \node[box, fill=gray!8, minimum height=0.75cm] (yaw) at (-2.75, -4.35) {\raisebox{-0.5\height}{\usebox{\iconyaw}}\hspace{3pt}\begin{tabular}{@{}c@{}}Yaw LQR\\(Sec.~\ref{sec:lqr})\end{tabular}};
  \node[box=3.0cm, fill=orange!4, minimum height=1.55cm] (alloc) at (2.05, -6.5)
    {Torque allocation \eqref{eq:torque_map}\\[2pt] $\vect\tau_w = \mat J^{-\top}\mat{\mathcal R}^\top\vect\tau_{\text{ball}}$};
  \node[box=2.75cm, fill=gray!8] (robot) at (-2.0, -6.5)
    {\raisebox{-0.5\height}{\includegraphics[height=1.35cm]{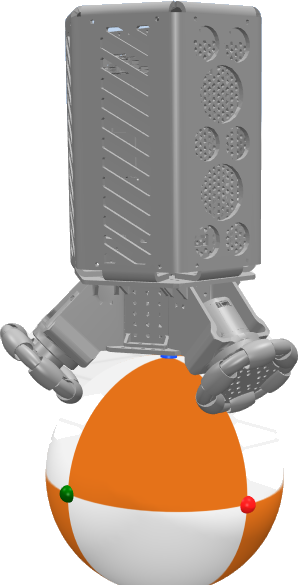}}\hspace{4pt}%
     \begin{tabular}{@{}c@{}}Ballbot\\(MuJoCo)\end{tabular}};
  \begin{scope}[on background layer]
    \node[layer=cyan, fit=(obs)(mpc),
          label={[font=\scriptsize\itshape, anchor=south west, inner sep=1pt]north west:Planning layer (every $\Delta t$)}] {};
    \node[layer=orange, fit=(eso)(smc)(cbf)(yaw),
          label={[font=\scriptsize\itshape, anchor=south west, inner sep=1pt]north west:Tracking layer (every $T_s$)}] {};
  \end{scope}
  \draw[meas, ->] (obs) -- node[below, black, font=\scriptsize, align=center] {$\hat{\vect p}_{i,k}$\\$q_k$} (mpc);
  \draw[meas, ->] ($(obs.west) + (-0.55, 0)$) -- node[above, black, font=\scriptsize] {$\hat{\vect c}_i$} (obs.west);
  \draw[ref, ->] (mpc.south -| smc) -- node[right, black] {$\vect p_r, \dot{\vect p}_r, \ddot{\vect p}_r$} (smc.north);
  \draw[est, ->] (eso) -- node[above, black] {$\hat{\vect q}, \hat{\vect f}$} (smc);
  \draw[trq, ->] (smc) -- node[above, black] {$u_{\text{HSMC}}$} (cbf);
  \draw[trq, ->] (cbf.south) -- node[pos=0.5, right, black, align=left] {$\tau_{\text{ball},x}$\\$\tau_{\text{ball},y}$} (cbf.south |- alloc.north);
  \coordinate (yawin) at ($(alloc.north west)!0.3!(alloc.north east)$);
  \draw[trq, ->] (yaw.east) -| node[pos=0.25, below, black] {$\tau_{\text{ball},z}$} (yawin);
  \draw[trq, ->] (alloc) -- node[above, black] {$\vect\tau_w$} (robot);
  \coordinate (bus) at (-4.15, 0);
  \draw[meas, ->] (robot.west) -- (bus |- robot.west) |- node[pos=0.8, above, black] {$\vect q$} (eso.west);
  \draw[meas, ->] (bus |- yaw) -- node[pos=0.35, above, black] {$\gamma$} (yaw.west);
  \fill[gray!60!black] (bus |- yaw) circle (1.5pt);
  \coordinate (ej) at ($(eso.east)+(0.28,0)$);
  \fill[violet] (ej) circle (1.5pt);
  \draw[est, ->] (ej) -- (ej |- 0,-3.82) -- (2.3, -3.82) node[midway, above=-1pt, black, xshift=22pt]
    {$\hat{\vect\varphi}, \hat{\dot{\vect\varphi}}, \hat{\vect f}_\varphi$} -- (2.3, -3.82 |- cbf.south);
  \draw[design, ->] (cbf.north) -- ++(0, 0.7) coordinate (dk) -| ($(mpc.south east)+(-0.4,0)$);
  \node[font=\scriptsize, anchor=north west, align=left, inner sep=2pt] at ($(dk)+(0.02,0.3)$) {$a_{\max} =$\\$\rho\varphi^\star/\kappa$};
  \node[font=\scriptsize, red!65!black, anchor=south west, inner sep=1pt] at ($(cbf.north)+(0.05,0.08)$) {$\varphi^\star$};
  \node[badge] at (smc.north east) {1};
  \node[badge] at (eso.north east) {1};
  \node[badge] at (cbf.north east) {1};
  \node[badge] at (mpc.north east) {2};
  \node[badge] at ($(dk)!0.5!(dk -| mpc.south east)$) {3};
  \node[badge] at (obs.north east) {4};
  \draw[sig, dashed, black!60, ->] (mpc.15) .. controls +(1.1, 0.4) and +(1.1, -0.4) ..
    node[right, black, align=left, font=\scriptsize] {plan\\continuity} (mpc.-15);
\end{tikzpicture}
\caption{\ifieee Two-layer architecture. References (cyan), torques (orange), measurements (gray), ESO estimates (violet); dashed red: acceleration bound \eqref{eq:amax}, set at design time from $\varphi^\star$. Badges: contributions of \cref{sec:intro}.\fi}
\label{fig:architecture}
\end{figure}

The navigation problem is split into two layers with separate roles (\cref{fig:architecture}). The planner generates a collision-free and tilt-feasible reference for the ball position; the tracker follows it, balances the body and enforces the tilt limit. The planner does not model the balance dynamics. It needs only three quantities from the lower layer: (i) the lean--acceleration relation \eqref{eq:lean}, which converts the tilt limit into a limit on the planned motion; (ii) the steady tilt $\varphi^\star$ that the tracker admits (\cref{prop:steady}), which fixes how much of the tilt limit the planner may use; and (iii) the robot footprint and a bound on the tracking error, which fix how far the plan must stay from the obstacles.

\paragraph{Why a reduced model} Planning on the full robot model, as in~\cite{pham2025cbf}, couples the planner to the balance dynamics: each solve integrates the nonlinear 3D model along the horizon, the problem changes with every change of the robot, and the tilt limit, which is handled by a separate filter, is not visible to the plan. Planning on a model of the ball alone removes all three issues, provided that the plan respects the limits of the balance. The model must therefore be simple enough to solve fast, rich enough to express those limits, and smooth enough for a tracker that turns accelerations into leans.

\paragraph{Time scales} The planner replans every $\Delta t$, much slower than the tracker period $T_s$; between replanning instants the tracker reads the current plan at its own rate. The two layers interact only through the reference and the three quantities above, so the planner can be changed without touching the tracker and vice versa.

\subsection{Jerk model and acceleration bound}

With the ball position, velocity and acceleration $\vect p, \vect v, \vect a \in \mathbb R^2$ and the jerk $\vect j$ held constant over a step $\Delta t$, the exact discretization of the triple integrator is
\begin{equation}
\begin{aligned}
  \vect p_{k+1} &= \vect p_k + \vect v_k\Delta t + \tfrac12\vect a_k\Delta t^2 + \tfrac16\vect j_k\Delta t^3,\\
  \vect v_{k+1} &= \vect v_k + \vect a_k\Delta t + \tfrac12\vect j_k\Delta t^2,\qquad
  \vect a_{k+1} = \vect a_k + \vect j_k\Delta t .
\end{aligned}
\label{eq:jerk}
\end{equation}
Between planner steps the reference is the cubic polynomial of \eqref{eq:jerk} evaluated at the current time. The two planes are independent in this model; they are coupled only through the norm bounds of the optimal control problem below.

The order of the model is chosen for the tracker. By \eqref{eq:lean}, the tracker leans the body by $\kappa\ddot{\vect p}_r$, so the smoothness of the lean reference is the smoothness of the planned acceleration.

\begin{remark}[Smooth lean reference]
\label{rem:smooth}
Because the jerk is held constant over each planner step, the planned acceleration changes linearly in time and never jumps, also across steps; the lean reference, which is proportional to it, is therefore continuous, and it changes at a rate bounded by the jerk bound times the lean gain. A lower-order model would lose this property: with the acceleration as the input, the planned acceleration, and with it the lean reference, would jump at every step, and the tracker would have to tilt the body instantly, which produces large torque peaks and triggers the tilt filter. The jerk model is thus the lowest order with a continuous lean reference, and it keeps the optimal control problem small. The rate of the lean reference still jumps at the planner steps\ifieee; the closed loop of the tracker smooths these corners\fi.
\end{remark}

\paragraph{Acceleration constraint}
\label{par:accel}
The planner bounds the acceleration so that the lean it asks for is one that the tracker can hold. By \eqref{eq:lean}, a planned acceleration $\vect a$ asks for the lean $\kappa\vect a$, and by \cref{prop:steady} the tracker holds a steady lean up to $\varphi^\star$ in one plane and up to $\bar\varphi^\star$ in the worst direction. The planner therefore uses
\begin{equation}
  \|\vect a\| \le a_{\max} = \frac{\rho\,\varphi^\star}{\kappa},\qquad
  0 < \rho \le \frac{\bar\varphi^\star}{\varphi^\star} ,
  \label{eq:amax}
\end{equation}
so that the requested lean stays below $\bar\varphi^\star$ in every direction, and in a single plane a margin of $(1-\rho)\varphi^\star$ remains for the corrections of the tracker. The bound depends on the tracker only through $\varphi^\star$ and on the robot only through $\kappa$, both in closed form. A bound derived from $\theta_{\max}$ instead asks for leans between $\varphi^\star$ and $\theta_{\max}$ that the tracker refuses; \cref{sec:ablation} quantifies the effect.

\subsection{Obstacle barrier functions}

Each obstacle $i$ is an ellipse with semi-axes $a_i \ge b_i$, center $\vect c_i(t)$ and orientation $\eta_i(t)$, moving with known velocity, acceleration and angular rate. In the obstacle frame, $\vect p^{E_i} = \mat{\mathcal R}(\eta_i)^\top(\vect p - \vect c_i)$, and the barrier is
\begin{equation}
  D_i(\vect p, t) = \left(\frac{p^{E_i}_x}{a_i + r}\right)^2 + \left(\frac{p^{E_i}_y}{b_i + r}\right)^2 - 1,
  \label{eq:obstacle_barrier}
\end{equation}
where $r = r_{\text{robot}} + r_{\text{track}}$, $r_{\text{robot}}$ is the footprint radius and $r_{\text{track}}$ a margin for the tracking error. The plan is safe if $D_i > 0$, i.e.\ the ball center stays outside the ellipse with semi-axes inflated by $r$.

\paragraph{Inflated ellipse}
\label{par:inflation}
Inflating the semi-axes is simpler than the exact offset of the ellipse by $r$, i.e.\ its Minkowski sum with a disc of radius $r$, which is not an ellipse. The inflated ellipse lies inside the exact offset, since in every direction its extent is at most that of the ellipse plus $r$; the two coincide for a circle, and the inflated ellipse falls short of the exact offset by at most
\begin{equation}
  r\,\frac{(\sqrt{a_i} - \sqrt{b_i})^2}{a_i + b_i},
  \label{eq:inflation_gap}
\end{equation}
which follows from the Kantorovich inequality and grows with the elongation of the ellipse. The barrier thus slightly underestimates the region that the footprint must avoid, and the margin $r_{\text{track}}$ must cover this gap as well as the tracking error.

\paragraph{Discrete-time CBF} Distance constraints at every step would force the plan to start the evasion only when an obstacle enters the horizon. The discrete-time CBF constraint~\cite{zeng2021}
\begin{equation}
  D_i(\vect p_{k+1}, t_{k+1}) \ge (1-\gamma)\,D_i(\vect p_k, t_k),\qquad 0 < \gamma < 1,
  \label{eq:dcbf}
\end{equation}
instead lets $D_i$ decrease by at most a fraction $\gamma$ per step, so the plan approaches an obstacle gradually; a small $\gamma$ makes the evasion start earlier and the plan more conservative. If the plan starts outside the inflated obstacle, $D_i(\vect p_0, t_0) > 0$, and satisfies \eqref{eq:dcbf} at every step, then by induction $D_i(\vect p_k, t_k) \ge (1-\gamma)^k D_i(\vect p_0, t_0) > 0$ for $k = 0, \dots, N$, so the plan is safe at the planner samples~\cite{zeng2021}. Between samples the plan is a cubic polynomial and the obstacle keeps moving; the step $\Delta t$ is short compared with the time the robot needs to cross the inflation $r$, and the margin $r_{\text{track}}$ also covers this inter-sample effect.

\paragraph{Concave obstacles} Following~\cite{pham2025cbf}, the polygon of a concave obstacle (in its own moving frame) is decomposed into convex pieces by an ear-clipping triangulation followed by Hertel--Mehlhorn merging~\cite{hertel1983}. Each piece is bounded by its minimum-area enclosing ellipse,
\begin{equation}
\begin{aligned}
  \min_{\vect c,\ \mat Q \succ 0}\ & -\log\det \mat Q\\
  \text{s.t.}\ & (\vect b_m - \vect c)^\top \mat Q\,(\vect b_m - \vect c) \le 1
  \quad \forall\ \text{vertices } \vect b_m ,
\end{aligned}
\end{equation}
solved with Khachiyan's algorithm~\cite{khachiyan1996}. All ellipses move rigidly with the obstacle, and each one contributes its own constraint \eqref{eq:dcbf}. The pieces cover the polygon much more tightly than one ellipse around the whole obstacle, which would close the opening of a concave shape; the price is one constraint per piece and per step.

\subsection{Optimal control problem and replanning}

At each replanning time $t_0$ the planner solves
\begin{equation}
\begin{aligned}
  \min_{\vect j_0,\dots,\vect j_{N-1}}\ & \sum_{k=1}^{N}
      \Bigl(w_p^{(k)}\|\vect p_k - \vect p_{\text{ref}}(t_k)\|^2
      + w_v\|\vect v_k - \vect v_{\text{ref}}(t_k)\|^2\\
  &\qquad + w_a\|\vect a_k\|^2 + w_j\|\vect j_{k-1}\|^2\Bigr)\\
  \text{s.t.}\ & \text{\eqref{eq:jerk}},\quad \vect z_0 = \vect z(t_0),\\
  & \|\vect v_k\| \le v_{\max},\quad \|\vect a_k\| \le a_{\max},\quad
      |\vect j_k|_\infty \le j_{\max},\\
  & D_i(\vect p_{k+1}, t_{k+1}) \ge (1-\gamma)\,D_i(\vect p_k, t_k)\quad \forall i, k ,
\end{aligned}
\label{eq:mpc}
\end{equation}
where $\vect z = [\vect p, \vect v, \vect a]$ and $\vect p_{\text{ref}}$, $\vect v_{\text{ref}}$ are the nominal path and its velocity.

\paragraph{Cost} The position and velocity terms pull the plan back to the nominal path once an obstacle has been passed; a larger terminal weight $w_p^{(N)}$ keeps the end of the horizon close to it. The acceleration term penalizes the lean that the plan asks from the tracker, and the jerk term penalizes the rate of change of that lean (\cref{rem:smooth}); together they make the plan prefer early, gentle evasions over late, sharp ones.

\paragraph{Structure} The dynamics \eqref{eq:jerk} are linear and the velocity, acceleration and jerk bounds are convex. The only nonconvex constraints are the barrier conditions \eqref{eq:dcbf}, one per obstacle ellipse and per step. The problem has six states and two inputs per stage, independent of the robot; its size grows linearly with the horizon and with the number of ellipses. It is solved with IPOPT~\cite{wachter2006} through CasADi~\cite{andersson2019}, warm-started with the shifted previous solution.

\begin{remark}[Plan continuity]
The initial state $\vect z(t_0)$ is the \emph{previous plan} evaluated at $t_0$, not the measured state. The reference is then continuous in position, velocity and acceleration across replanning instants, so the lean reference stays continuous as well (\cref{rem:smooth}), and the planner does not react to the tracking error, which is bounded by the margin $r_{\text{track}}$.
\end{remark}

\begin{remark}[Feasibility]
With moving obstacles, \eqref{eq:mpc} is not guaranteed to remain feasible from one replanning instant to the next: an obstacle may enter the horizon too fast for the bounded acceleration to avoid it. The constraints of \eqref{eq:mpc} are kept hard for known obstacle motion; with predicted obstacles, \cref{sec:conformal} softens them and adds a fallback plan. \cref{sec:sim} reports the number of solver failures.
\end{remark}

\section{Uncertainty-Aware Planning with Conformal Prediction}
\label{sec:conformal}

\cref{sec:main} assumes that the future motion of every obstacle is known. In practice the planner only observes the obstacles and must predict where they will be over its horizon; the prediction is wrong whenever an obstacle turns or changes its speed. This section replaces the known motion by a prediction and bounds the prediction error with conformal prediction~\cite{vovk2005,angelopoulos2021}, following the idea of~\cite{lindemann2023conformal} (\cref{fig:conformal}): the obstacle ellipses of \eqref{eq:obstacle_barrier} are inflated at every horizon step by a radius that the true obstacle position does not exceed with a prescribed probability. The planner, the tracker and the interface of \cref{sec:main} are unchanged.

\subsection{Obstacle prediction}

At every replanning instant $t_0$ the planner has noisy observations $\hat{\vect c}_i(t_0 - j\Delta t)$, $j = 0, \dots, H-1$, of the center of obstacle~$i$. A predictor maps them to the predicted centers $\hat{\vect p}_{i,k|t_0}$ at $t_k = t_0 + k\Delta t$, $k = 0, \dots, N$. Any predictor can be used; the simplest one fits a constant velocity by least squares,
\begin{equation}
\begin{gathered}
  (\hat{\vect c}_0, \hat{\vect v}) = \arg\min_{\vect c, \vect v}
  \sum_{j=0}^{H-1} \bigl\|\hat{\vect c}_i(t_0 - j\Delta t) - \vect c + j\Delta t\,\vect v\bigr\|^2,\\
  \hat{\vect p}_{i,k|t_0} = \hat{\vect c}_0 + k\Delta t\,\hat{\vect v} .
\end{gathered}
\label{eq:cv_predictor}
\end{equation}
The prediction error grows with $k$, since the obstacle may turn or change its speed within the horizon; its distribution is unknown.

\begin{figure}[!t]
\centering
\begin{tikzpicture}[>={Stealth[length=4pt, width=3pt]}, font=\footnotesize]
  \begin{scope}
    \draw[->] (0, 0) -- (3.3, 0) node[right] {$S_k$};
    \draw[->] (0, 0) -- (0, 2.0);
    \foreach \x/\h in {0.1/0.55, 0.3/1.2, 0.5/1.6, 0.7/1.5, 0.9/1.2, 1.1/0.95, 1.3/0.72,
                       1.5/0.55, 1.7/0.42, 1.9/0.3, 2.1/0.22, 2.3/0.15, 2.5/0.1, 2.7/0.07, 2.9/0.05} {
      \pgfmathsetmacro{\c}{\x < 2.15 ? 1 : 0}
      \ifdim\c pt=1pt
        \fill[cyan!45, draw=cyan!60!black, line width=0.3pt] (\x - 0.09, 0) rectangle (\x + 0.09, \h);
      \else
        \fill[red!35, draw=red!60!black, line width=0.3pt] (\x - 0.09, 0) rectangle (\x + 0.09, \h);
      \fi
    }
    \draw[dashed, red!70!black, line width=0.8pt] (2.2, 0) -- (2.2, 1.75) node[above] {$q_k$};
    \node[cyan!50!black, align=center, font=\scriptsize] at (1.0, 1.85) {$\ge 1-\alpha$ of the\\calibration scores};
    \node[font=\scriptsize\itshape] at (1.6, -0.55) {(a) offline calibration, step $k$};
  \end{scope}
  \begin{scope}[xshift=3.55cm, yshift=-0.35cm]
    \draw[densely dashed, gray] (2.4, -0.15) -- (2.4, 2.5) node[above, font=\scriptsize] {nominal};
    \foreach \x in {0.15, 0.4, 0.65} \fill[gray!70!black] (\x, 1.0) circle (0.035);
    \fill[gray!45] (0.9, 1.0) circle (0.2);
    \node[font=\scriptsize] at (0.75, 1.42) {$\hat{\vect c}_i(t_0)$};
    \draw[densely dashed, gray!70!black] (0.9, 1.0) -- (3.7, 1.0);
    \foreach \k/\x/\R/\o in {1/1.9/0.36/0.55, 2/2.8/0.52/0.45, 3/3.7/0.68/0.35} {
      \fill[red!30, fill opacity=\o] (\x, 1.0) circle (\R);
      \draw[red!60!black, densely dotted] (\x, 1.0) circle (\R);
      \fill[gray!70!black] (\x, 1.0) circle (0.035);
      \node[red!60!black, font=\scriptsize] at (\x, {1.0 - \R - 0.13}) {$k{=}\k$};
    }
    \draw[blue!65!black, line width=0.8pt] (0.9, 1.0) .. controls (1.9, 1.08) and (2.9, 1.3) .. (3.65, 1.55);
    \node[blue!65!black, font=\scriptsize] at (3.95, 1.85) {true};
    \fill[black] (2.4, -0.15) circle (0.06);
    \node[font=\scriptsize, right] at (2.45, -0.2) {robot};
    \draw[cyan!70!black, line width=1.1pt, ->] plot[smooth, tension=0.7] coordinates
          {(2.4, -0.15) (2.0, 0.2) (1.55, 0.55) (1.22, 1.2) (1.4, 1.8) (1.9, 2.22) (2.35, 2.4)};
    \foreach \k/\x/\y in {1/2.0/0.2, 2/1.55/0.55, 3/1.22/1.2} {
      \fill[cyan!70!black] (\x, \y) circle (0.05);
      \node[cyan!50!black, font=\scriptsize, left] at (\x, \y) {$\k$};
    }
    \node[cyan!50!black, font=\scriptsize] at (0.95, 2.05) {plan};
    \node[font=\scriptsize\itshape] at (2.0, -0.65) {(b) online planning at $t_0$};
  \end{scope}
\end{tikzpicture}
\caption{\ifieee Conformal prediction. (a) Offline, $q_k$ is the quantile covering a fraction $1-\alpha$ of the calibration errors $S_k$. (b) Online, the predicted obstacle at step $k$ is inflated by $r + q_k$, and the plan (cyan) stays outside.\fi}
\label{fig:conformal}
\end{figure}

\subsection{Conformal prediction regions}

For horizon step $k$, the nonconformity score is the prediction error
\begin{equation}
  S_k = \bigl\|\vect c_i(t_k) - \hat{\vect p}_{i,k|t_0}\bigr\| .
  \label{eq:score}
\end{equation}
A calibration set of $n$ scores $S_k^{(1)}, \dots, S_k^{(n)}$ is collected from independent recorded obstacle trajectories (or from a simulator of the obstacles), and the conformal radius is the empirical quantile
\begin{equation}
  q_k = S_k^{(\lceil (n+1)(1-\alpha) \rceil)} ,
  \label{eq:conformal_radius}
\end{equation}
the $\lceil (n+1)(1-\alpha)\rceil$-th smallest calibration score, with a miscoverage level $\alpha \in (0, 1)$.

The radius inherits the coverage guarantee of split conformal prediction~\cite{vovk2005,angelopoulos2021}: if the score $S_k$ of a new prediction and the calibration scores are exchangeable and $\alpha \ge 1/(n+1)$, then
\begin{equation}
  \mathbb P\bigl(S_k \le q_k\bigr) \ \ge\ 1 - \alpha .
  \label{eq:coverage}
\end{equation}
The argument uses only that, by exchangeability, the rank of $S_k$ among the $n+1$ scores is uniformly distributed; the condition on $\alpha$ makes the index in \eqref{eq:conformal_radius} at most $n$. The guarantee holds for any predictor and any distribution of the obstacle motion, but only per horizon step. A guarantee for the whole horizon follows from the union bound by calibrating every step at the level $\alpha/N$, at the price of larger radii.

\subsection{Uncertainty-aware obstacle barriers}

The barrier \eqref{eq:obstacle_barrier} of step $k$ is evaluated at the predicted center with the semi-axes inflated by $q_k$,
\begin{equation}
  D_{i,k}(\vect p) = \left(\frac{p^{E_i}_x}{a_i + r + q_k}\right)^2
                   + \left(\frac{p^{E_i}_y}{b_i + r + q_k}\right)^2 - 1,
  \label{eq:conformal_barrier}
\end{equation}
with $\vect p^{E_i} = \mat{\mathcal R}(\eta_i)^\top(\vect p - \hat{\vect p}_{i,k|t_0})$, and replaces $D_i(\vect p_k, t_k)$ in \eqref{eq:dcbf} and \eqref{eq:mpc}.

\paragraph{Probabilistic clearance} The inflation by $q_k$ turns the coverage of the prediction into a clearance of the plan, as in~\cite{lindemann2023conformal}. For a disc obstacle, $a_i = b_i$, the constraint $D_{i,k}(\vect p_k) \ge 0$ keeps the planned position at the distance $a_i + r + q_k$ from the predicted center. With probability at least $1-\alpha$ the true center lies within $q_k$ of the predicted one by \eqref{eq:coverage}, and then, by the triangle inequality,
\begin{equation}
  \bigl\|\vect p_k - \vect c_i(t_k)\bigr\| \ \ge\ \bigl\|\vect p_k - \hat{\vect p}_{i,k|t_0}\bigr\| - S_k \ \ge\ a_i + r ,
  \label{eq:prob_clearance}
\end{equation}
i.e.\ the planned footprint, enlarged by the tracking margin, does not intersect the true obstacle at step $k$. For an elongated ellipse, inflating the semi-axes by $q_k$ is smaller than the offset of the ellipse by $q_k$, and the clearance holds up to the gap \eqref{eq:inflation_gap} evaluated with $r + q_k$ instead of $r$. The tracking error is covered by $r_{\text{track}}$ as in \cref{sec:main}.

\subsection{Feasibility: soft barriers and fallback plan}

The conformal radii grow along the horizon, so the inflated region of an obstacle expands from one step to the next. In \eqref{eq:dcbf} this expansion decreases $D_{i,k}$ even when the planned position does not move toward the obstacle; together with the change of the prediction between replanning instants, the problem can become infeasible. Two mechanisms keep the planner well defined.

\paragraph{Soft barrier constraints} The barrier constraints are relaxed by nonnegative slacks,
\begin{equation}
  D_{i,k+1}(\vect p_{k+1}) \ \ge\ (1-\gamma)\,D_{i,k}(\vect p_k) - \sigma_{i,k},
  \qquad \sigma_{i,k} \ge 0,
  \label{eq:soft_dcbf}
\end{equation}
and $w_\sigma\sum_{i,k}\sigma_{i,k}$ with a large weight $w_\sigma$ is added to the cost of \eqref{eq:mpc}. The problem is then always feasible, and a nonzero slack signals that the hard constraints could not be met; the number of such replanning instants is reported.

\paragraph{Fallback plan} If the solver does not converge, the planner keeps the previous plan, with its original time origin, instead of applying an unconverged iterate. The previous plan satisfied the constraints for the prediction at its own replanning instant; the next solve is warm-started from it.

\begin{remark}[Assumptions and limits]
\label{rem:conformal}
The coverage \eqref{eq:coverage} requires the calibration data to be representative of the deployment, i.e.\ exchangeable with it; a change of the obstacle behavior (for example a different crowd density) calls for recalibration or for adaptive variants of conformal prediction. The guarantee concerns the plan at the planner samples; the tracking error and the motion between samples are covered by the margin $r_{\text{track}}$, as in \cref{sec:main}. Finally, a better predictor does not change the guarantee but reduces $q_k$ and therefore the conservatism of the plan.
\end{remark}

\section{Simulation Results}
\label{sec:sim}

All results are obtained in MuJoCo~\cite{todorov2012} using the detailed asRoBallet base model adapted from the authors' released MJCF implementation~\cite{wan2026asroballet} (free base, three omni wheels with 36 passive rollers, ball--floor contact with friction), at $2$\,ms per step with the controllers at the same rate and the wheel torques saturated at $\pm 4$\,N\,m.

The experiments proceed from the components to the complete scheme. \Cref{sec:sim_models} checks that the planar and 3D design models reproduce MuJoCo and determines up to which tilt the planes decouple. \Cref{sec:sim_tracker} evaluates whether the HSMC-ESO-ECBF keeps the tilt limit on a step and a push despite the input-gain mismatch, and \cref{sec:sim_steady} checks the steady tilt that the robust ECBF admits (\cref{prop:steady}). The complete scheme is then applied to static, moving, rotating and concave obstacles in \cref{sec:sim_mpc}; \cref{sec:ablation} isolates the contribution of the tracker-consistent acceleration bound and of the tracking margin, \cref{sec:sim_nmpc} compares the scheme with a full-model NMPC followed by a tilt filter, and \cref{sec:robust} tests its robustness to random obstacles and to model mismatch.{} Finally, \cref{sec:sim_conformal} evaluates the conformal radii against obstacles whose motion is predicted.

\subsection{Simulation setup}
\label{sec:setup}

\paragraph{Robot} The ball has radius $R = 0.115$\,m, mass $m_b = 3.021$\,kg and inertia $I_b = 0.0245$\,kg\,m$^2$. The three omni wheels of radius $r_w = 0.050$\,m are spaced $120^\circ$ apart with axles tilted $45^\circ$ from the vertical, each with twelve passive rollers and an axle torque of at most $\pm 4$\,N\,m (gear $2$, control range $\pm 2$). The body (base only) has mass $M = 5.642$\,kg and inertia $\mathrm{diag}(0.080, 0.080, 0.035)$\,kg\,m$^2$, and its COM lies $l = 0.241$\,m above the ball center, $0.63^\circ$ ahead of the base axis. These values are taken from the adapted asRoBallet MJCF model~\cite{wan2026asroballet}. The kinematic matrix \eqref{eq:J} is
\begin{equation*}
  \mat J = \begin{bmatrix}
    -1.626 & 0 & 1.626\\
     0.813 & -1.409 & 1.626\\
     0.813 & 1.409 & 1.626
  \end{bmatrix},
\end{equation*}
with singular values $\{2.82, 1.99, 1.99\}$ (condition number $\sqrt 2$); with the body fixed, MuJoCo reproduces it within $1$--$4\,\%$. The linearized plane \eqref{eq:linear} has $a_1 = -64.83$, $a_2 = 57.59$\,s$^{-2}$, $b_1 = 17.505$, $b_2 = -9.164$\,(kg\,m$^2$)$^{-1}$, i.e.\ unstable poles $\pm 7.59$\,s$^{-1}$ (time constant $0.13$\,s), and the lean gain \eqref{eq:lean} is $\kappa = 0.193$\,rad per m/s$^2$ ($11.0^\circ$ per m/s$^2$).

\paragraph{Tracker} The tracker runs every $T_s = 2$\,ms (500\,Hz). The ESO has $\varepsilon = 0.01$ and $(\gamma_1, \gamma_2, \gamma_3) = (3, 3, 1)$, a triple pole at $-1/\varepsilon$. The sliding surface places the poles of \cref{prop:surface} at $2, 3, 4$\,s$^{-1}$; with $\kappa_0 = -414.0$ and $\beta = -1$ this gives the positive gains $(k_\alpha, k_1, k_\beta, k_2) = (0.063, 0.923, 0.229, 4.77)$, and the reaching law uses $k = 15$, $\eta = 0.5$ and $\Phi = 0.05$. The ECBF uses $\theta_{\max} = 0.1$\,rad ($5.73^\circ$), $\ell_1 = \ell_2 = 10$ and $\delta = 0.5$\,rad/s$^2$, for which \cref{prop:steady} gives $\varphi^\star = 3.54^\circ$ in one plane and $\bar\varphi^\star = 2.97^\circ$ in the worst direction, so $\bar\varphi^\star/\varphi^\star = 0.84$. The yaw LQR uses $\mat Q = \mathrm{diag}(20, 1)$, which gives $\mat K = [-4.33,\ -1.11]$. The LQR baseline uses $\mat Q = \mathrm{diag}(30, 200, 10, 10)$, $R_u = 1$: $\mat K = [-5.09,\ -88.83,\ -5.22,\ -17.53]$, closed-loop poles $\{-1.77, -4.51 \pm 0.15j, -63.0\}$\,s$^{-1}$.

\paragraph{Planner} The MPC replans every $\Delta t = 0.1$\,s (10\,Hz) over $N = 25$ steps ($2.5$\,s), with the bounds $v_{\max} = 0.5$\,m/s and $j_{\max} = 2$\,m/s$^3$, the cost weights $(w_p, w_p^{(N)}, w_v, w_a, w_j) = (10, 50, 1, 1, 0.05)$ and the DCBF rate $\gamma = 0.2$. With $\rho = 0.8 \le \bar\varphi^\star/\varphi^\star$, \eqref{eq:amax} gives $a_{\max} = 0.26$\,m/s$^2$; the bound from $\theta_{\max}$ used in the ablation is $0.7\,\theta_{\max}/\kappa = 0.36$\,m/s$^2$. The footprint radius is $r_{\text{robot}} = 0.25$\,m and the tracking margin $r_{\text{track}} = 0.10$\,m. With $r = r_{\text{robot}} + r_{\text{track}} = 0.35$\,m, the gap \eqref{eq:inflation_gap} between the inflated ellipses and the exact offset of the obstacles is zero for the circles and at most $2.0$\,cm for the translating ellipses and $4.0$\,cm for the pieces of the U-shaped obstacle (bound \eqref{eq:inflation_gap}: $7.5$\,cm), well inside $r_{\text{track}}$. The solver converged at every replanning instant in the four scenarios, the Monte Carlo runs and the model-mismatch runs.

\paragraph{Conformal calibration} The calibration data are $200$ independent trajectories of $30$\,s of the obstacles of \cref{sec:sim_conformal}, each started at a random position and heading; on each, the predictor \eqref{eq:cv_predictor} is applied at $20$ random instants, which gives $n = 4000$ score vectors $(S_0, \dots, S_N)$. The radii $q_k$ of \eqref{eq:conformal_radius} are computed once, offline, for every horizon step with $\alpha = 0.1$, and the planner uses them as a fixed table. On an independent test set generated in the same way, the per-step coverage is $0.893$--$0.910$ (target $0.90$; mean $0.900$). The radius grows from $1.7$\,cm at $k = 0$ to $18$\,cm at $1$\,s and $61$\,cm at the end of the horizon ($2.5$\,s). The scores of one trajectory are correlated, so the calibration set is only approximately exchangeable with a new prediction; the test coverage confirms the level. The obstacles of Scenario~4 in \cref{sec:sim_conformal} stay inside an arena, which changes their motion, so their radii are calibrated separately in the same way: the per-step coverage is $0.897$--$0.908$ (mean $0.904$), and the radius grows from $1.7$\,cm to $21$\,cm at $1$\,s and $72$\,cm at $2.5$\,s.

\subsection{Model validation}
\label{sec:sim_models}

\begin{figure}[!t]
\centering
\ifieeeplots\includegraphics[width=\figw]{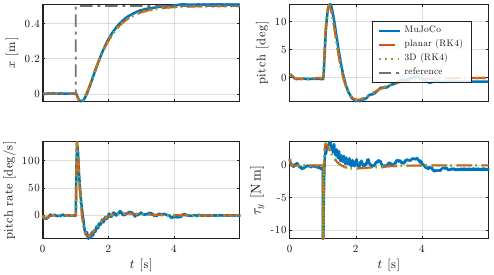}\fi
\caption{LQR on MuJoCo, the planar and the 3D model: step $x_r: 0 \to 0.5$\,m.}
\label{fig:compare_step}
\end{figure}

\begin{figure}[!t]
\centering
\includegraphics[width=\figw]{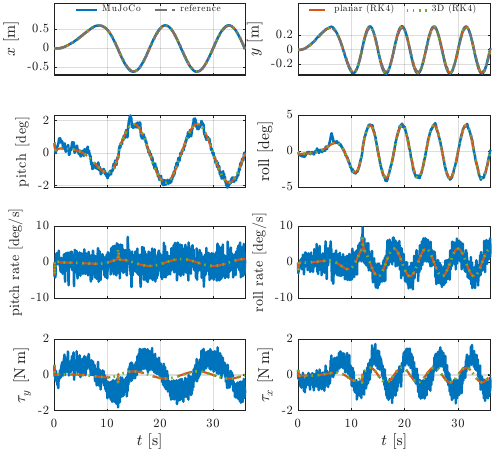}
\caption{Figure eight ($A = 0.6$\,m, $12$\,s per lap): $x$--$z$ plane (left) and $y$--$z$ plane (right).}
\label{fig:compare_figure8}
\end{figure}

The LQR runs with identical gains, references and saturation on MuJoCo, on the planar model \eqref{eq:nonlinear} and on the 3D model \eqref{eq:kane}, both integrated with RK4 from the measured initial MuJoCo state (\cref{tab:model_nominal}, \cref{fig:compare_step,fig:compare_figure8}).

\begin{table}[!t]
\centering
\begin{threeparttable}
\caption{RK4 models vs.\ MuJoCo.}
\label{tab:model_nominal}
\footnotesize\setlength{\tabcolsep}{2.5pt}\renewcommand{\arraystretch}{1.15}
\begin{tabular*}{\linewidth}{@{\extracolsep{\fill}}l*{4}{S[table-format=2.1]S[table-format=1.2]}@{}}
\toprule
& \multicolumn{2}{c}{\hd{Tilt $5^\circ$}} & \multicolumn{2}{c}{\hd{Step $0.5$\,m}} & \multicolumn{2}{c}{\hd{Sine}} & \multicolumn{2}{c}{\hd{Figure eight}} \\
\cmidrule(lr){2-3}\cmidrule(lr){4-5}\cmidrule(lr){6-7}\cmidrule(l){8-9}
{\hd{Signal}} & {\hd{RMS}} & {\hd{$r$}} & {\hd{RMS}} & {\hd{$r$}} & {\hd{RMS}} & {\hd{$r$}} & {\hd{RMS}} & {\hd{$r$}} \\
\midrule
$x$          & 15.7 & 0.94 & 1.7 & 1.00 & 2.7 & 1.00 & 1.8 & 1.00 \\
pitch        & 7.9 & 0.86 & 3.5 & 0.99 & 4.4 & 1.00 & 8.0 & 0.99 \\
$\dot x$     & 7.0 & 0.91 & 3.0 & 1.00 & 3.3 & 1.00 & 3.5 & 1.00 \\
pitch rate   & 3.6 & 0.97 & 2.2 & 0.99 & 16.3 & 0.84 & 28.1 & \color{badred}0.30 \\
$\tau_y$     & 4.3 & 0.68 & 6.3 & 0.77 & 35.2 & \color{badred}0.45 & 35.5 & \color{badred}0.28 \\
\bottomrule
\end{tabular*}
\begin{tablenotes}[flushleft]
\footnotesize
\item[] RMS difference is expressed in \% of the MuJoCo peak; $r$ is the correlation coefficient. Planar and 3D results are equal to the shown precision.
\end{tablenotes}
\end{threeparttable}
\end{table}

The positions and tilts follow MuJoCo within a few percent; the rates and torques differ more on the slow references, where Coulomb friction of the wheels and rollers, absent from both models, is comparable to the control torque. The planar and 3D models are equal to $10^{-3}$\,degrees here.

\begin{figure}[!t]
\centering
\ifieeeplots\includegraphics[width=\columnwidth]{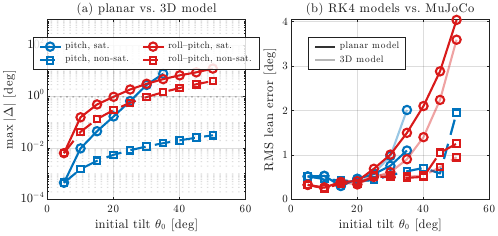}\fi
\caption{(a) Largest planar--3D difference versus initial tilt (dotted: $1^\circ$). (b) RMS lean error of each RK4 model with respect to MuJoCo.}
\label{fig:sweep}
\end{figure}

\paragraph{Validity of the decoupled planes} The robot is released from an initial tilt $\theta_0 = 5^\circ \dots 50^\circ$ about $y$ (pitch only) or about $(1, 1, 0)/\sqrt2$ (combined roll--pitch), with and without the wheel saturation (\ifieee\cref{fig:sweep}\fi).

A pitch-only lean stays planar without saturation (difference $< 0.04^\circ$ up to $50^\circ$). A combined roll--pitch lean couples the planes: the difference exceeds $1^\circ$ at $\theta_0 \approx 20^\circ$ with saturation and $\approx 30^\circ$ without it. Saturation amplifies the coupling, because clipping each wheel separately turns the ball torque vector and creates a yaw torque; the 3D model is then closer to MuJoCo than the planar one ($1.41^\circ$ vs.\ $2.11^\circ$ RMS at $40^\circ$). With saturation and a pitch-only lean of $40^\circ$, the 3D model falls while MuJoCo and the planar model recover, so the 3D model is not uniformly more accurate. The planar design models are adequate up to $\approx 15^\circ$ of combined tilt; the tilt limit of $5.73^\circ$ keeps the robot well inside this range.

\subsection{Tracking controller}
\label{sec:sim_tracker}

HSMC-ESO with and without ECBF, and LQR, are compared on a step of the target to $x = 1$\,m and a $25$\,N push on the body for $0.1$\,s (\cref{tab:hsmc_results}, \cref{fig:hsmc}).

\begin{table}[!t]
\centering
\begin{threeparttable}
\caption{Tracker comparison in MuJoCo.}
\label{tab:hsmc_results}
\footnotesize\setlength{\tabcolsep}{4pt}\renewcommand{\arraystretch}{1.15}
\begin{tabular*}{\linewidth}{@{\extracolsep{\fill}}lccc@{}}
\toprule
\hd{Metric} & \hd{HSMC-ESO-ECBF} & \hd{HSMC-ESO} & \hd{LQR} \\
\midrule
\tabgroup{4}{Step of $1$\,m}
max tilt [deg]            & $\mathbf{5.71}$ & \bad{$21.16$} & \bad{$25.42$} \\
undershoot [cm]           & $\mathbf{0.2}$ & $6.1$ & $8.1$ \\
settling to $\pm5$\,cm [s] & $2.91$ & $2.24$ & $\mathbf{2.03}$ \\
control variation [1/s]   & $10.5$ & $\mathbf{5.5}$ & $16.5$ \\
ECBF active [\%]          & $34.6$ & -- & -- \\
\midrule
\tabgroup{4}{Push of $25$\,N for $0.1$\,s}
max tilt [deg]            & $\mathbf{4.50}$ & \bad{$11.21$} & \bad{$10.76$} \\
max displacement [cm]     & $27.2$ & $13.6$ & $\mathbf{12.9}$ \\
control variation [1/s]   & $9.8$ & $\mathbf{3.8}$ & $9.7$ \\
\bottomrule
\end{tabular*}
\begin{tablenotes}[flushleft]
\footnotesize
\item[] $\theta_{\max} = 5.73^\circ$; red: tilt limit exceeded. Control variation: $\frac1T\sum_k\|\text{ctrl}_{k+1}-\text{ctrl}_k\|_1$. ECBF active: share of time the filter modifies the command.
\end{tablenotes}
\end{threeparttable}
\end{table}

\ifieeeplots
\begin{figure}[!t]
\centering
\includegraphics[width=\figw]{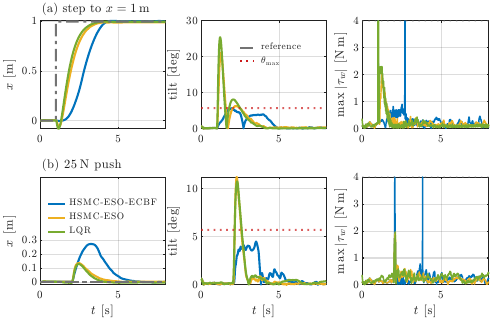}
\caption{\ifieee (a) Step to $x = 1$\,m and (b) $25$\,N push: the ECBF keeps the tilt below $5.73^\circ$, at the price of a longer braking distance.\else (a) Step to $x = 1$\,m: without the ECBF, HSMC-ESO and LQR lean up to $21$--$25^\circ$; with it the tilt stays below $5.73^\circ$. (b) $25$\,N push: the ECBF limits the braking lean, so the robot travels twice as far ($0.27$ vs.\ $0.14$\,m) before returning.\fi}
\label{fig:hsmc}
\end{figure}
\fi

\paragraph{Input-gain mismatch and robust term} Stepping two copies of the same balanced state with and without a ball-torque pulse shows that MuJoCo produces only $\approx 70\,\%$ of the modeled tilt acceleration ($-6.2$ vs.\ $-9.2$\,rad/s$^2$ per N\,m over $2$--$4$\,ms): the wheels and rollers absorb part of the torque and the ball reacts through compliant roller contacts. The ECBF relies on the predicted tilt acceleration, so this mismatch, together with the observer lag, lets the tilt exceed the limit unless the observer is fast enough and the robust term covers the remaining error (\cref{tab:ecbf_delta}). A constant margin is not usable: it makes the condition infeasible at the upright posture for tight limits.

\begin{table}[!t]
\centering
\begin{threeparttable}
\caption{Overshoot of the tilt limit, $\max\theta - \theta_{\max}$ [deg].}
\label{tab:ecbf_delta}
\footnotesize\setlength{\tabcolsep}{3pt}\renewcommand{\arraystretch}{1.15}
\begin{tabular*}{\linewidth}{@{\extracolsep{\fill}}S[table-format=1.2]S[table-format=1.1]*{4}{S[table-format=+1.2,retain-explicit-plus]}S[table-format=1.1]@{}}
\toprule
& & \multicolumn{2}{c}{\hd{$\theta_{\max} = 3^\circ$}} & \multicolumn{2}{c}{\hd{$\theta_{\max} = 5.73^\circ$}} & {\hd{Fig.\ eight}} \\
\cmidrule(lr){3-4}\cmidrule(lr){5-6}
{\hd{$\varepsilon$}} & {\hd{$\delta$}} & {\hd{step}} & {\hd{push}} & {\hd{step}} & {\hd{push}} & {\hd{RMS [cm]}} \\
\midrule
 0.02 & 0 & \color{badred}+4.75 & \color{badred}+4.73 & \color{badred}+4.50 & \color{badred}+2.74 & 1.9 \\
 0.01 & 0 & \color{badred}+1.24 & \color{badred}+1.40 & \color{badred}+1.34 & \color{badred}+0.54 & 2.5 \\
 0.01 & 0.2 & \color{badred}+0.82 & \color{badred}+1.01 & \color{badred}+1.04 & \color{okgreen}-0.17 & 2.5 \\
\rowcolor{rowhl}0.01 & 0.5 & +0.01$^\ast$ & \color{okgreen}-0.40 & \color{okgreen}-0.02 & \color{okgreen}-1.23 & 3.9 \\
\bottomrule
\end{tabular*}
\begin{tablenotes}[flushleft]
\footnotesize
\item[] Red: limit violated; green: limit respected. $^\ast$ Target not reached in $8$\,s. Shaded: chosen setting.
\end{tablenotes}
\end{threeparttable}
\end{table}

The ECBF holds the limit at the price of speed, because accelerating and braking a ballbot require leaning: a tighter limit gives a slower step and a longer braking distance after a push. A reference that asks for leans close to $\varphi^\star$ therefore cannot be tracked accurately with the filter active, which is what the acceleration bound of the planner avoids.

\subsection{Steady tilt admitted by the robust ECBF}
\label{sec:sim_steady}

\begin{figure}[!t]
\centering
\includegraphics[width=\figw]{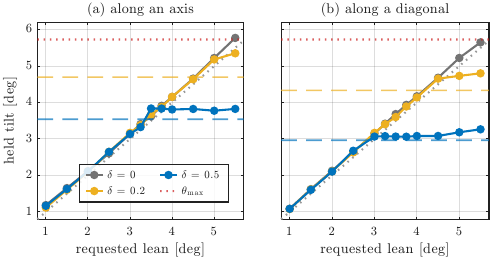}
\caption{\ifieee Steady tilt held in MuJoCo versus the requested lean, (a) along an axis and (b) along a diagonal; dashed: $\varphi^\star$ and $\bar\varphi^\star$ (\cref{prop:steady}).\fi}
\label{fig:steady_tilt}
\end{figure}

To check \cref{prop:steady}, the tracker alone follows a reference with a constant acceleration $a$, which asks for the constant lean $\varphi_{\text{ref}} = \kappa a$ by \eqref{eq:lean}, along the $x$ axis or along a diagonal. The acceleration rises smoothly within $1$\,s and is then held for $4$\,s; the held tilt is the mean measured tilt over the last $1.5$\,s. \cref{fig:steady_tilt} shows it for $\delta = 0$, $0.2$ and $0.5$, with the other parameters of \cref{sec:setup}.

Without the robust term, the held tilt follows the request over the whole range, and at the largest request ($5.5^\circ$) it exceeds $\theta_{\max}$ by $0.04^\circ$. With $\delta > 0$ it saturates, lower along the diagonal than along the axis and lower for the larger $\delta$, as \cref{prop:steady} predicts. For the chosen $\delta = 0.5$, the held tilt saturates at $3.8^\circ$ along the axis and at $3.1^\circ$ along the diagonal, $0.3^\circ$ and $0.1^\circ$ above $\varphi^\star = 3.54^\circ$ and $\bar\varphi^\star = 2.97^\circ$; for $\delta = 0.2$ the excess is up to $0.65^\circ$. Below the saturation the held tilt exceeds the request by $0.1$--$0.2^\circ$. The excess is consistent with the input-gain mismatch above: at the saturation the filter corrects the input for a tilt acceleration that MuJoCo produces only in part, so the body settles at a larger tilt than in the design model. The planner requests at most $\rho\varphi^\star = 2.83^\circ$, below the saturation in both directions.

\subsection{Jerk-model MPC-CBF obstacle avoidance}
\label{sec:sim_mpc}

The complete scheme drives the robot from $(0, 0)$ to $(4, 4)$\,m along a nominal line at $0.3$\,m/s, in four scenarios following~\cite{pham2025cbf}: \emph{static} (two circles), \emph{moving} (two circles crossing the path at $0.26$\,m/s, timed to meet the robot at $t = 7$ and $13$\,s), \emph{ellipse} (two $0.5 \times 0.18$\,m ellipses that translate and rotate at $\pm 0.5$\,rad/s), and \emph{concave} (the U-shaped obstacle bounded by three ellipses, accelerating across the path and turning at $0.2$\,rad/s). For the U ($1.2 \times 0.9$\,m, area $0.63$\,m$^2$) the three ellipses cover $1.02$\,m$^2$, against $1.70$\,m$^2$ for one ellipse around the U. The clearance is the distance from the ball center to the obstacle boundary minus the robot radius. The paths are shown in \cref{fig:mpc_xy} and the time histories in \cref{fig:mpc_ts}. Time-lapses of the four runs in MuJoCo are shown in \cref{fig:hook}(c)--(f).

\begin{table}[!t]
\centering
\begin{threeparttable}
\caption{Jerk-model MPC-CBF planning with HSMC-ESO-ECBF tracking in MuJoCo.}
\label{tab:mpc_results}
\footnotesize\setlength{\tabcolsep}{2pt}\renewcommand{\arraystretch}{1.1}
\begin{tabular*}{\linewidth}{@{\extracolsep{\fill}}cccc@{}}
\toprule
\hd{Static} & \hd{Moving} & \hd{Ellipse} & \hd{Concave} \\
\midrule
\tabgroup{4}{Min clearance [cm], robot / plan}
$11.8$ / $10.0$ & $11.1$ / $10.0$ & $10.3$ / $10.1$ & $11.0$ / $10.0$ \\
\tabgroup{4}{Max plan deviation [m] / max tilt [deg]}
$0.52$ / $4.06$ & $0.74$ / $4.07$ & $0.66$ / $4.05$ & $1.15$ / $4.23$ \\
\midrule
\tabgroup{4}{Tracking error [cm], RMS / max / final}
$2.0$ / $3.8$ / $0.9$ & $2.0$ / $4.5$ / $0.1$ & $1.9$ / $3.8$ / $0.4$ & $1.6$ / $3.5$ / $0.8$ \\
\midrule
\tabgroup{4}{MPC solve time [ms], median / 95th percentile / max$^\ast$}
$10$ / $16$ / $17$ & $12$ / $16$ / $21$ & $9$ / $15$ / $17$ & $11$ / $18$ / $21$ \\
\bottomrule
\end{tabular*}
\begin{tablenotes}[flushleft]
\footnotesize
\item[] $\theta_{\max} = 5.73^\circ$, $\varphi^\star = 3.54^\circ$, $r_{\text{track}} = 10$\,cm; arrival at $18.8$\,s in all scenarios. $^\ast$ Excluding the first (cold-start) solve at $t = 0$, which takes $64$--$114$\,ms.
\end{tablenotes}
\end{threeparttable}
\end{table}

\ifieeeplots
\begin{figure}[!t]
\centering
\includegraphics[width=\columnwidth]{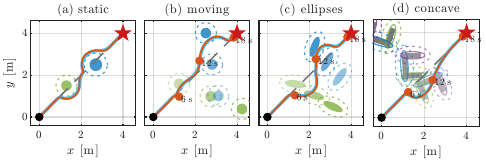}
\caption{\ifieee Paths in the four scenarios: nominal line ({\color{refgray}gray dashed}), MPC-CBF plan ({\color{plancyan}cyan}) and ball path in MuJoCo ({\color{ballorange}orange}), from the start (black dot) to the goal ({\color{limitred}red star}). Obstacles are filled and their inflated boundaries dotted; moving obstacles are drawn at $t = 6$, $12$, $18$\,s (darker = later), with the robot (dots) at the same times.\else Paths in the four scenarios: nominal line, MPC-CBF plan and MuJoCo ball path. Obstacles are filled and their footprint-inflated boundaries dotted; moving obstacles are drawn at $t = 6$, $12$ and $18$\,s (darker = later) with the robot (dots) at the same times. In (d) the U-shaped obstacle (gray) is bounded by three ellipses.\fi}
\label{fig:mpc_xy}
\end{figure}
\fi

\ifieeeplots
\begin{figure}[!t]
\centering
\includegraphics[width=\figw]{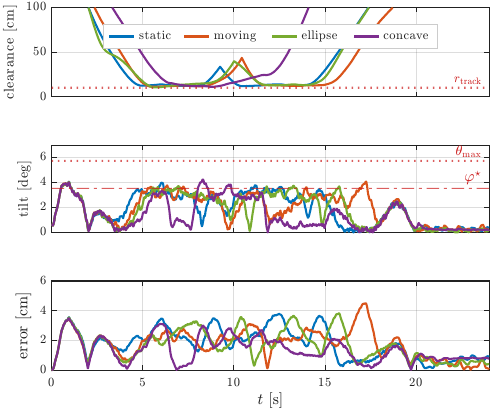}
\caption{Time histories of the four scenarios: smallest clearance of the robot to the obstacles, measured tilt, and tracking error.}
\label{fig:mpc_ts}
\end{figure}
\fi

In all scenarios the robot keeps more than $10$\,cm between its footprint and the obstacles, the tilt stays below $\theta_{\max}$, the robot arrives at the same time as on the nominal line ($18.8$\,s; the detours are absorbed by the velocity margin), and the MPC solves in $9$--$12$\,ms (median) and at most $21$\,ms, well below the $100$\,ms replanning period. Only the first solve, started without a warm start while the robot is at rest, takes $64$--$114$\,ms. The measured tilt follows the lean implied by the planned acceleration; the excess over $\rho\varphi^\star = 2.8^\circ$, up to $1.4^\circ$, comes from tracking corrections.

\subsection{Ablation: acceleration bound and tracking margin}
\label{sec:ablation}

\cref{tab:ablation} compares the proposed planner with two variants: an acceleration bound derived from the tilt limit ($a_{\max} = 0.7\,\theta_{\max}/\kappa = 0.36$\,m/s$^2$) instead of $\varphi^\star$, and no tracking margin.

\begin{table*}[!t]
\centering
\begin{threeparttable}
\caption{Ablation of the planner.}
\label{tab:ablation}
\footnotesize\setlength{\tabcolsep}{4.5pt}\renewcommand{\arraystretch}{1.15}
\begin{tabular*}{\linewidth}{@{\extracolsep{\fill}}l*{4}{S[table-format=2.1]}*{4}{S[table-format=2.1]}*{4}{S[table-format=2.1]}c@{}}
\toprule
& \multicolumn{4}{c}{\hd{Min clearance [cm]}} & \multicolumn{4}{c}{\hd{RMS error [cm]}} & \multicolumn{4}{c}{\hd{Max error [cm]}} & {\hd{Goal}} \\
\cmidrule(lr){2-5}\cmidrule(lr){6-9}\cmidrule(lr){10-13}
{\hd{Variant}} & {\hd{S}} & {\hd{M}} & {\hd{E}} & {\hd{C}} & {\hd{S}} & {\hd{M}} & {\hd{E}} & {\hd{C}} & {\hd{S}} & {\hd{M}} & {\hd{E}} & {\hd{C}} & {\hd{reached}} \\
\midrule
\rowcolor{rowhl}Proposed: $a_{\max} = \rho\varphi^\star/\kappa$, $r_{\text{track}} = 10$\,cm
  & 11.8 & 11.1 & 10.3 & 11.0 & 2.0 & 2.0 & 1.9 & 1.6 & 3.8 & 4.5 & 3.8 & 3.5 & $4/4$ \\
$a_{\max}$ from $\theta_{\max}$
  & 11.2 & \color{badred}6.8 & 10.7 & 11.1 & 6.7 & 3.6 & 5.4 & \color{badred}11.7 & \color{badred}18.5 & 10.0 & \color{badred}16.7 & \color{badred}39.1 & $3/4$ \\
No tracking margin ($r_{\text{track}} = 0$)
  & \color{badred}1.6 & \color{badred}1.4 & \color{badred}0.7 & \color{badred}1.9 & 1.8 & 2.0 & 2.2 & 1.5 & 3.5 & 4.5 & 4.8 & 3.5 & $4/4$ \\
\bottomrule
\end{tabular*}
\begin{tablenotes}[flushleft]
\footnotesize
\item[] S: static; M: moving; E: ellipse; C: concave. Red: clearance below $r_{\text{track}} = 10$\,cm or tracking error above $10$\,cm. Shaded: proposed.
\end{tablenotes}
\end{threeparttable}
\end{table*}

With the bound derived from $\theta_{\max}$, the plan asks for leans between $\varphi^\star$ and $0.7\,\theta_{\max}$ that the ECBF refuses; the robot lags the plan by up to $39$\,cm, the clearance in the moving scenario drops to $6.8$\,cm, and in the concave scenario the robot does not reach the goal within $24$\,s. The proposed bound reduces the RMS tracking error by a factor of $1.8$--$7$. Without the tracking margin, the plan touches the footprint-inflated boundary and the robot keeps only $0.7$--$1.9$\,cm of clearance.

\subsection{Comparison with a full-model NMPC}
\label{sec:sim_nmpc}

As a baseline, the full-model approach of~\cite{pham2025cbf} is applied to the same robot and scenarios: a nonlinear MPC on the 3D model of \cref{sec:model3d} computes the ball torques directly, with the same inflated ellipses and discrete-time CBFs for the obstacles, and the tilt limit is removed from the MPC and enforced afterwards by the same robust ECBF filter. As in~\cite{pham2025cbf}, the prediction horizon is $1$\,s ($20$ steps of $0.05$\,s) and the control horizon $0.5$\,s; the barrier decay per second equals that of the jerk-model planner, and the barrier constraints are softened by a heavily penalized slack, without which the problem became infeasible as soon as an obstacle entered the horizon. The NMPC is re-solved every $20$\,ms from the measured state, and the simulation waits for each solve, which favors the baseline: on a real robot the first input would be applied only after the solve time.

\begin{table}[!t]
\centering
\begin{threeparttable}
\caption{Jerk-model MPC-CBF (proposed) versus full-model NMPC-CBF with a tilt filter.}
\label{tab:nmpc}
\footnotesize\setlength{\tabcolsep}{0pt}\renewcommand{\arraystretch}{1.15}
\begin{tabular*}{\linewidth}{@{\extracolsep{\fill}}l*{2}{S[table-format=-2.1]}*{2}{S[table-format=1.2]}*{2}{S[table-format=3.0]}@{}}
\toprule
 & \multicolumn{2}{c}{\hd{Min clear. [cm]}} & \multicolumn{2}{c}{\hd{Max tilt [deg]}} & \multicolumn{2}{c}{\hd{Solve, median [ms]}} \\
\cmidrule(l{4pt}r{4pt}){2-3}\cmidrule(l{4pt}r{4pt}){4-5}\cmidrule(l{4pt}){6-7}
{\hd{Scenario}} & {\hd{Prop.}} & {\hd{NMPC}} & {\hd{Prop.}} & {\hd{NMPC}} & {\hd{Prop.}} & {\hd{NMPC}} \\
\midrule
static  & 11.8 & \color{badred}-11.2 & 4.06 & 5.21 & 10 & 145 \\
moving  & 11.1 & \color{badred}-12.3 & 4.07 & \color{badred}8.02 & 12 & 131 \\
ellipse & 10.3 & \color{badred}-2.3 & 4.05 & 5.38 & 9 & 150 \\
concave & 11.0 & \color{badred}-3.7 & 4.23 & \color{badred}6.30 & 11 & 108 \\
\bottomrule
\end{tabular*}
\begin{tablenotes}[flushleft]
\footnotesize
\item[] Prop.: proposed. Red: footprint inside the inflated obstacle or tilt above $\theta_{\max} = 5.73^\circ$.
\end{tablenotes}
\end{threeparttable}
\end{table}

\cref{tab:nmpc} summarizes the results. The full-model NMPC enters the inflated region of an obstacle in all four scenarios, by up to $12$\,cm, and exceeds the tilt limit in two of them, while the proposed scheme keeps more than $10$\,cm of clearance and a tilt below $4.3^\circ$. The cause is the inconsistency between the plan and the tilt filter discussed in \cref{sec:related}: the NMPC does not know the tilt limit and plans evasions that require large leans; the filter modifies its input during $69$--$86\,\%$ of the time (against $26$--$41\,\%$ for the proposed scheme), the robot does not execute the evasion that the NMPC planned, and the next solve starts from a state closer to the obstacle. Two variants confirm this. Without the tilt filter, the NMPC keeps $10.5$\,cm of clearance in the static scenario but leans the body by up to $23^\circ$, four times the limit. With the same $2.5$\,s horizon as the jerk-model planner, it still enters the inflated obstacle ($-4.9$\,cm), leans by up to $8.0^\circ$, and stops $53$\,cm short of the goal.

\ifieee The full model is also costly: the median solve time of the NMPC is $108$--$150$\,ms, $10$--$15$ times that of the jerk-model planner and above the $20$\,ms period at which it must run to balance the robot.\fi

\subsection{Robustness: random scenarios and model mismatch}
\label{sec:robust}

\paragraph{Random scenarios} In $50$ Monte Carlo runs the robot drives along the same nominal line among two random obstacles: a circle (radius $0.20$--$0.35$\,m) or a rotating ellipse (semi-axes $0.30$--$0.50$ by $0.12$--$0.25$\,m, rate up to $\pm0.5$\,rad/s), moving at up to $0.3$\,m/s in a random direction and crossing the line within $\pm0.3$\,m of the robot at a random time between $4$ and $16$\,s, at least $4$\,s apart. \ifieee In all runs there is no collision, no fall and no solver failure, the goal is reached, the tilt stays below $4.7^\circ$, and the RMS tracking error is $1.8$--$3.4$\,cm (median $2.4$\,cm), with a final error of at most $1.7$\,cm; the median minimum clearance is $11.1$\,cm.\fi In $11$ runs the clearance falls below the margin $r_{\text{track}}$ (at worst $3.1$\,cm), mostly with an obstacle moving at $0.2$--$0.3$\,m/s: the tracking error during the evasive maneuver (up to $12$\,cm) then uses part of the margin.

\paragraph{Model mismatch} The MuJoCo model is changed after the planner and the tracker are designed from the nominal model: the body mass and inertia by $\pm20\,\%$, and the height of the body COM by $\pm20\,\%$ ($\pm4.8$\,cm). The tilt is measured as on a real robot, from the base orientation and the nominal COM offset. A heavier or lighter body and a higher COM leave the results almost unchanged; the ESO absorbs the model error. A lower COM is the hardest case: the body leans more for the same acceleration than the planner assumes ($\kappa$ grows), the tracking error reaches $10.2$\,cm and the clearance $6.0$\,cm, still without collision and with the tilt below the limit.\ifieee{} In the worst case over the four scenarios, the tilt stays below $4.4^\circ$ and the final error below $1$\,cm for all models; the mass changes keep the clearance at $9.8$--$10.5$\,cm and the RMS error at $2.0$--$2.5$\,cm, against $10.9$ and $2.1$\,cm with the nominal model.\fi

\subsection{Predicted obstacles with conformal prediction}
\label{sec:sim_conformal}

\ifieeeplots
\begin{figure}[!t]
\centering
\includegraphics[width=\columnwidth]{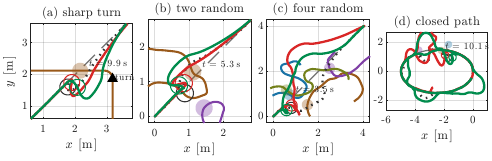}
\caption{Predicted obstacles, paths: (a) Scenario 1 (triangle: start of the turn), (b) one run of Scenario 2, (c) Scenario 3, (d) Scenario 4, first $90$\,s. Robot with the oracle ({\color{black}black dotted}), the constant-velocity prediction CV ({\color{cvred}red}) and CV with conformal inflation ({\color{cpgreen}green}); obstacle paths (except in (d)) and footprints at the smallest clearance of the CV planner.}
\label{fig:conformal_turn}
\end{figure}

\begin{figure}[!t]
\centering
\includegraphics[width=\columnwidth]{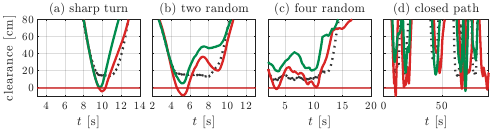}
\caption{True clearance between the footprint and the nearest obstacle in the runs of \cref{fig:conformal_turn}, same colors.}
\label{fig:conformal_clearance}
\end{figure}
\fi

\paragraph{Setup} The obstacles are discs of radius $0.25$\,m whose speed and turn rate follow Ornstein--Uhlenbeck processes~\cite{uhlenbeck1930} (mean speed $0.25$\,m/s, speed limited to $0.45$\,m/s), so that they turn and change speed unpredictably. The planner observes the obstacle centers every $0.1$\,s with Gaussian noise of $1$\,cm, predicts them with \eqref{eq:cv_predictor} from the last $H = 5$ observations, inflates them by the radii of \cref{sec:setup}, and softens the barriers as in \eqref{eq:soft_dcbf} with $w_\sigma = 10^4$. In the random scenarios (Scenarios~2 and~3), each obstacle heads for the nominal line of the robot and would meet it at a random time; scenarios in which an obstacle occupies the start or the goal are discarded.

\begin{table}[!t]
\centering
\begin{threeparttable}
\caption{Predicted obstacles, Scenario~2: $30$ random scenarios per planner, soft barriers.}
\label{tab:conformal}
\footnotesize\setlength{\tabcolsep}{0pt}\renewcommand{\arraystretch}{1.15}
\begin{tabular*}{\linewidth}{@{\extracolsep{\fill}}lS[table-format=1.0]S[table-format=-1.1]S[table-format=2.1]S[table-format=1.2]S[table-format=3.0]S[table-format=2.0]@{}}
\toprule
 & & \multicolumn{2}{c}{\hd{Clearance [cm]}} & & \multicolumn{2}{c}{\hd{Solver}} \\
\cmidrule(lr){3-4}\cmidrule(l){6-7}
{\hd{Planner}} & {\hd{Coll.}} & {\hd{min}} & {\hd{median}} & {\hd{Tilt [deg]}} & {\hd{slack}} & {\hd{[ms]}} \\
\midrule
oracle (true future)        & 0 & 6.9 & 13.1 & 4.60 & 0 & 18 \\
CV                          & \color{badred}7 & \color{badred}-8.8 & 6.5 & 4.81 & 392 & 18 \\
\rowcolor{rowhl}CV + CP     & 0 & 5.2 & 28.0 & 4.57 & 329 & 22 \\
\bottomrule
\end{tabular*}
\begin{tablenotes}[flushleft]
\footnotesize
\item[] CV: constant-velocity prediction \eqref{eq:cv_predictor}; CP: conformal
inflation \eqref{eq:conformal_barrier}. Coll.: runs with a collision;
clearance: minimum over each run; tilt: maximum over all runs; slack:
replanning instants (of $7200$) with a nonzero slack in \eqref{eq:soft_dcbf};
[ms]: median solve time. Shaded: proposed.
\end{tablenotes}
\end{threeparttable}
\end{table}

\paragraph{Scenario 1: an obstacle that turns sharply} A single obstacle moves north at $0.4$\,m/s, parallel to the robot's path and clear of it, then turns left by $90^\circ$ at $t = 7$\,s within $1$\,s and crosses the nominal line where the robot arrives at $t \approx 10$\,s. Before the turn, the constant-velocity prediction places the obstacle away from the path. \cref{fig:conformal_turn}(a)\ifieeeplots{} and \cref{fig:conformal_clearance}(a)\fi{} compare the three planners, all with soft barriers. With the constant-velocity prediction the footprint enters the obstacle by $4.1$\,cm. With the conformal inflation the regions of the later horizon steps already cover the turn, the plan starts to evade earlier, and the robot passes with $0.8$\,cm of clearance (oracle: $14.1$\,cm).
Even sharper or later turns (for example at $0.5$\,m/s, or within $0.8$\,s) defeat both predictors: such maneuvers lie outside the calibration data, which violates the exchangeability assumption of \eqref{eq:coverage} (\cref{rem:conformal}).

\paragraph{Scenario 2: two random obstacles} \cref{tab:conformal} compares three planners on $30$ scenarios with two random obstacles each, all with the soft barriers \eqref{eq:soft_dcbf}; \cref{fig:conformal_turn}(b) shows one of them, in which the constant-velocity planner enters an obstacle by $8.8$\,cm while the conformal planner keeps $5.2$\,cm (oracle: $12.8$\,cm). The oracle, which knows the future obstacle positions, never collides. The constant-velocity prediction without inflation collides in $7$ of $30$ runs, by up to $9$\,cm, whenever an obstacle turns toward the planned path. With the conformal inflation no run collides; the price is a larger detour (median clearance $28$\,cm against $13$\,cm for the oracle), since the radii at the end of the horizon are large. With predicted obstacles a nonzero slack is needed at about $5\,\%$ of the replanning instants; with hard barrier constraints the solver would fail at these instants instead (between $3$ and $6\,\%$ of them in the same scenarios), mostly when the inflated region of an obstacle expands faster than \eqref{eq:dcbf} allows. All solves take a median of $18$--$22$\,ms, well below the $100$\,ms period.

\begin{table}[!t]
\centering
\begin{threeparttable}
\caption{Predicted obstacles, Scenarios~3 and~4: one run per planner, soft barriers.}
\label{tab:conformal_runs}
\footnotesize\setlength{\tabcolsep}{0pt}\renewcommand{\arraystretch}{1.15}
\begin{tabular*}{\linewidth}{@{\extracolsep{\fill}}lS[table-format=1.0]S[table-format=-1.1]S[table-format=1.2]S[table-format=3.0]c@{}}
\toprule
{\hd{Planner}} & {\hd{Coll.}} & {\hd{Clearance [cm]}} & {\hd{Tilt [deg]}} & {\hd{slack}} & {\hd{Laps}} \\
\midrule
\multicolumn{6}{@{}l}{\emph{Scenario 3: four random obstacles}} \\
oracle (true future)        & 0 & 9.0 & 4.46 & 0 & -- \\
CV                          & \color{badred}1 & \color{badred}-1.4 & 4.57 & 37 & -- \\
\rowcolor{rowhl}CV + CP     & 0 & 19.4 & 4.33 & 10 & -- \\
\midrule
\multicolumn{6}{@{}l}{\emph{Scenario 4: closed path, $90$\,s}} \\
oracle (true future)        & 0 & 11.7 & 4.51 & 0 & 1.97 \\
CV                          & \color{badred}3 & \color{badred}-9.4 & 4.64 & 97 & 1.72 \\
\rowcolor{rowhl}CV + CP     & \color{badred}1 & \color{badred}-3.7 & 4.61 & 100 & 1.96 \\
\bottomrule
\end{tabular*}
\begin{tablenotes}[flushleft]
\footnotesize
\item[] Coll.: number of times the footprint enters an obstacle, exits shorter than $0.5$\,s not counted; clearance: minimum over the run; tilt: maximum over the run; slack: replanning instants (of $240$ in Scenario~3, of $900$ in Scenario~4) with a nonzero slack in \eqref{eq:soft_dcbf}; laps: completed laps of the closed path. Shaded: proposed.
\end{tablenotes}
\end{threeparttable}
\end{table}

\paragraph{Scenario 3: four random obstacles} Four obstacles of the same random motion, so that the radii remain valid, cross the nominal line one after the other within the windows $4$--$6$, $6.5$--$8.5$, $9$--$11$ and $11.5$--$13.5$\,s. \cref{fig:conformal_turn}(c) and \cref{tab:conformal_runs} show one run. With the constant-velocity prediction the footprint enters an obstacle by $1.4$\,cm; with the conformal inflation the clearance stays above $19.4$\,cm (oracle: $9.0$\,cm). Both planners end within $3$\,cm of the goal at $t = 20$\,s.

\paragraph{Scenario 4: endless run on a closed path} The robot follows an elliptical path of semi-axes $2.5$ and $1.5$\,m at $0.3$\,m/s, without end, while three obstacles with the motion above roam across it and steer back whenever they leave a box of half-size $3.0 \times 1.9$\,m around the center of the path. A reference parametrized by time would make a robot that fell behind chase a point that has moved on. The reference is therefore parametrized by the progress along the path: at every replanning instant, the planned position is projected on the path, searching only up to $1.5$\,m ahead of the last progress, and the reference over the horizon starts at this point at the path speed. The progress never decreases, so that the robot continues from where it is after a detour instead of turning back. \cref{fig:conformal_turn}(d) and \cref{tab:conformal_runs} show the first $90$\,s of one run. The oracle does not collide. With the constant-velocity prediction the footprint enters an obstacle three times, by up to $9.4$\,cm, and the robot completes $1.72$ laps. With the conformal inflation the robot completes $1.96$ laps, about as many as the oracle, and the footprint enters an obstacle once, by $3.7$\,cm for about $0.6$\,s at $t \approx 70$\,s. The inflation thus reduces the collisions but does not remove them: \eqref{eq:coverage} holds per horizon step and per obstacle, not for a whole run, and a long run with several obstacles leaves more occasions for the true motion to leave the regions. A guarantee over a run would need the smaller level of the union bound and larger radii, which is not evaluated here.

\section{Conclusion}
\label{sec:conclusion}

A two-layer navigation scheme for a three-omni-wheel ballbot was presented: a jerk-model MPC with discrete-time CBFs plans collision-free references for the ball, and an HSMC-ESO-ECBF tracker balances the robot and enforces the tilt cone. The layers are linked by an acceleration bound derived from the largest steady tilt that the robust ECBF admits, which proved essential: a bound derived from the tilt limit alone multiplied the tracking error by up to $7$ and caused a mission failure. On a contact-rich MuJoCo model with wheel saturation, the scheme avoided static, moving, rotating and concave obstacles with more than $10$\,cm of clearance, tracked the plan within $2$\,cm RMS and planned in about $10$\,ms per step. A full-model NMPC with the same obstacle constraints and a separate tilt filter, in contrast, entered the inflated obstacles in all four scenarios and needed $10$--$15$ times more computation. In $50$ random scenarios the scheme reached the goal without collision every time, and it tolerated a $\pm 20\,\%$ change of the body mass and of the COM height. When the obstacle motion is only predicted, inflating the obstacles by conformal prediction radii removed the collisions of a constant-velocity predictor (from $7$ to $0$ of $30$ random runs) and handled a turning obstacle that the predictor alone hit, while soft barrier constraints kept every planning problem feasible; maneuvers outside the calibration data remain a limit. In one run of $90$\,s of an endless path-following task on a closed path among three roaming obstacles, the inflation reduced the collisions from three to one, but did not remove them, since the coverage holds per step and not over a run. Open points are replanning from the measured state, learned obstacle predictors and recalibration when the obstacle behavior changes, a time-varying robust bound from the ESO error, the reflected wheel inertia in the input gain, the wheel saturation inside the ECBF quadratic program, and validation on hardware.

\bibliographystyle{IEEEtran}
\bibliography{references}

\end{document}